\documentclass{article}
\usepackage{ijcai26}

\usepackage{times}
\usepackage{soul}
\usepackage{url}
\usepackage[hidelinks]{hyperref}
\usepackage[utf8]{inputenc}
\usepackage[small]{caption}
\usepackage{graphicx}
\usepackage{amsmath}
\usepackage{amsthm}
\usepackage{booktabs}
\usepackage{algorithm}
\usepackage{algpseudocode}
\usepackage[switch]{lineno}
\usepackage{microtype}
\usepackage{graphicx}
\usepackage{subcaption}
\usepackage{booktabs} 
\usepackage[dvipsnames]{xcolor}
\usepackage{makecell}
\usepackage{amssymb} 

\usepackage{amsmath}
\usepackage{amssymb}
\usepackage{mathtools}

\usepackage{amsthm}
\usepackage{booktabs,longtable,siunitx,adjustbox,lscape}
\usepackage{graphicx}
\usepackage{multirow}
\usepackage{caption}
\usepackage{array}
\usepackage{subcaption}
\usepackage{float} 
\usepackage{kotex}
\usepackage{booktabs, makecell}
\usepackage{xcolor}
\usepackage{tabularx}
\usepackage{caption}
\usepackage{pifont}
\usepackage{enumitem}
\usepackage{placeins} 
\usepackage{subfiles}
\usepackage{makecell}
\usepackage{booktabs}   
\usepackage{tabularx}   
\usepackage{array}      
\usepackage{makecell}   
\usepackage{colortbl}
\usepackage{xcolor}
\usepackage[table]{xcolor}

\newcommand{\R}{{\mathbb R}}
\newcommand{\xv}{{\mathbf x}}

\newcommand{\sv}{{\mathbf s}}

\newcommand{\gv}{{\mathbf g}}

\newcommand{\Sc}{{\mathcal S}}
\newcommand{\Tc}{{\mathcal T}}

\newcommand{\data}{{\mathcal D}}

\theoremstyle{definition}
\newtheorem*{definition3.1}{Definition 3.1}
\newcommand{\defeq}{\overset{\text{\tiny def}}{=}}
\newcommand{\ldefeq}{\mathrel{\raisebox{-0.3ex}{$\defeq$}}}

\newcommand\footnotetextcopyrightpermission[1]{}

\newcommand{\cmark}{\ding{51}} 
\newcommand{\xmark}{\ding{55}} 

\newcolumntype{Y}{>{\centering\arraybackslash}X}

\usepackage[capitalize,noabbrev]{cleveref}

\theoremstyle{plain}
\newtheorem{theorem}{Theorem}[section]

\theoremstyle{definition}
\newtheorem{definition}[theorem]{Definition}

\theoremstyle{remark}

\usepackage[textsize=tiny]{todonotes}

\title{INSHAPE: Instance-Level Shapelets for Interpretable Time-Series Classification}

\author{
Seongjun Lee$^{1\ast}$\and
Seokhyun Lee$^{1\ast}$\And
Changhee Lee$^{1\dagger}$\\
\affiliations
$^1$Department of Artificial Intelligence, Korea University\\
\emails
\{pyoung7307, nshuhsn, changheelee\}@korea.ac.kr
}

\begin{document}

\maketitle
\renewcommand{\thefootnote}{}
\footnotetext{$^\ast$ Equal contribution.}
\footnotetext{$^\dagger$ Corresponding author.}
\renewcommand{\thefootnote}{\arabic{footnote}}

\begin{abstract}
Discovering \textit{shapelets} -- i.e.,  discriminative temporal patterns within time series -- has been widely studied to address the inherent complexity of time-series classification (TSC) and to make model decision-making processes more transparent. 
However, existing methods primarily focus on \textit{population-level} shapelets optimized across the entire dataset, which leads to two fundamental limitations: (i) population-level patterns often misalign with instance-specific features, resulting in suboptimal performance and potentially misleading interpretations, and (ii) most methods treat shapelets as independent entities, overlooking important temporal dependencies and interactions among multiple patterns.
To address these limitations, we propose \textit{INSHAPE}, an interpretable TSC framework that discovers variable-length, discriminative temporal patterns specific to each time series.
INSHAPE identifies these patterns as non-overlapping segments and models their temporal dependencies, thereby providing clear instance-level interpretations while achieving strong predictive performance. 
Furthermore, INSHAPE bridges local and global interpretability through a bottom-up approach, aggregating instance-level shapelets into prototypical (population-level) shapelets. 
Extensive experiments on 128 UCR and 30 UEA benchmark datasets show that INSHAPE consistently outperforms state-of-the-art shapelet-based methods while providing more intuitive and interpretable insights. 
\end{abstract}

\section{Introduction}

\begin{figure}[!t]
    \centering
    \includegraphics[width=\linewidth]{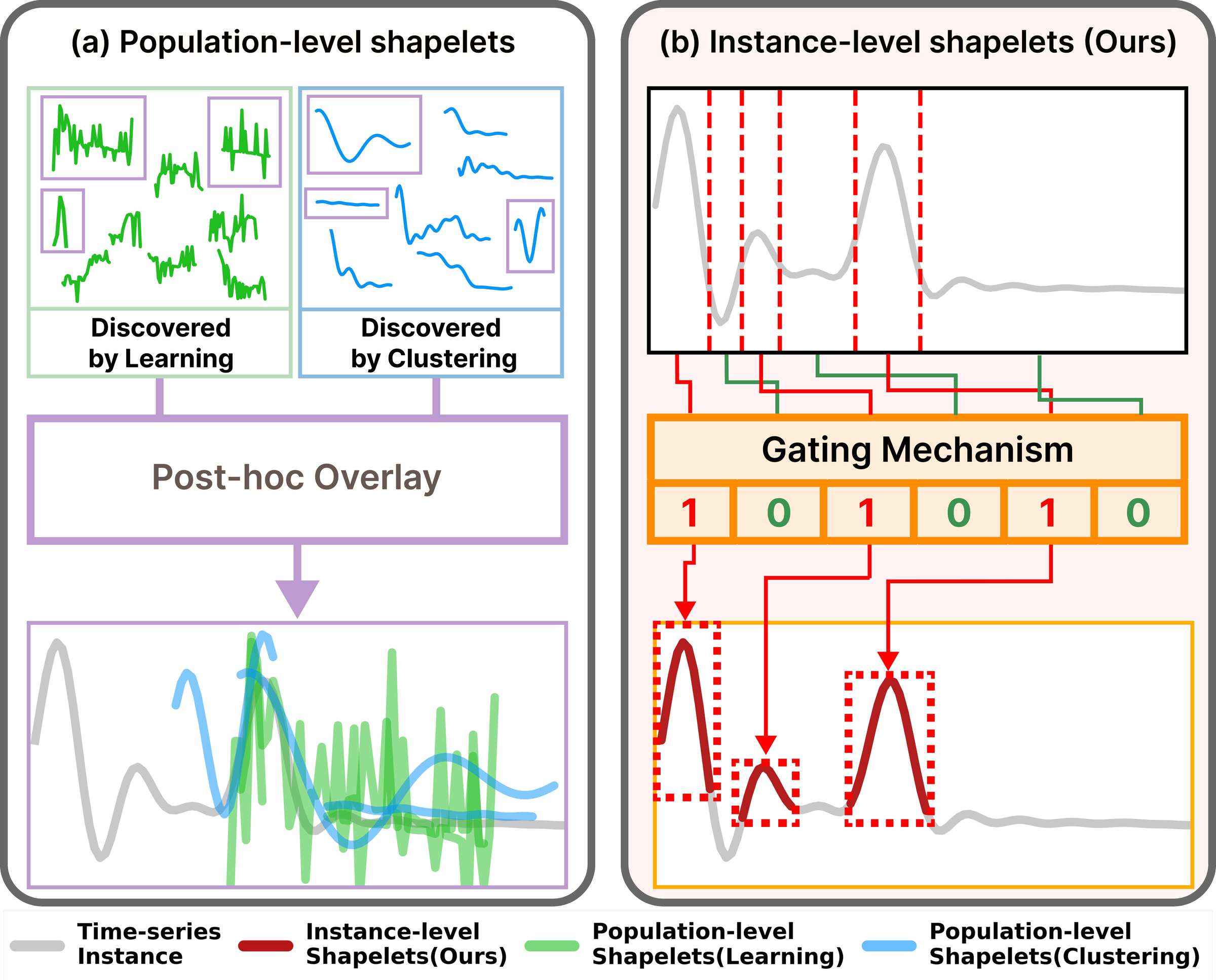}
    \caption{
    \textbf{Local interpretation with (a) population-level shapelets and (b) instance-level shapelets.}
    Population-level shapelets (i.e., SBM, ShapeNet) often result in overlapping and less interpretable patterns, whereas our instance-level shapelets show non-overlapping discriminative regions that align well with the input time series.
    }
    \label{fig1:instance_vs_population}
\end{figure}

Shapelets are highly informative subsequences that distinguish between classes in time-series data~\cite{koegh_Shapelet_KDD2009}. By discovering class-discriminative temporal patterns across an entire population, shapelets provide \textit{global} interpretability, revealing which temporal features differentiate one class from another. In addition, shapelets also support \textit{local} interpretability by pinpointing whether and where these patterns occur within individual instances -- typically through post-hoc overlay. These two levels of interpretability are particularly valuable in high-stakes domains such as healthcare~\cite{hyland_early_NatureMedicine2020,gumbsch_enhancingbiomarker_Bioinformatics2020,li_patientwarning_Heliyon2023,qin_fetalet_ComplexIntelligent2025,neves_shapeletECG_ComputersinBiologyandMedicine}: global explanations help clinicians identify population-level temporal biomarkers, while local explanations provide case-specific insights for understanding individual-level diagnoses.

While recent deep learning models for time-series classification (TSC) achieve strong predictive performance, their inherent complexity has driven the development of shapelet-based methods that integrate interpretable temporal patterns directly into the prediction process~\cite{li_shapenet_AAAI2021,zuo2_SVPT_AAAI2023,qu_ShapeConv_ICLR2024,wen_InterpGN_ICLR2025,liu_SoftShape_ICML2025}.
However, these methods face two critical limitations when applied to instance-level prediction and interpretation.
First, most methods rely on \textit{population-level shapelets} optimized to capture broad class characteristics, which often result in patterns that are overly smoothed or poorly aligned with individual instances. 
Second, despite the dependence of TSC decisions on the interaction, order, or relative position of multiple temporal patterns~\cite{mohammadi_deepTSCsurvey2024}, existing methods typically discover shapelets independently, and therefore fail to capture these crucial temporal dependencies.
Figure~\ref{fig1:instance_vs_population}(a) illustrates the consequence of these limitations: when overlapped and input-misaligned population-level shapelets are presented as local explanations, they obscure which temporal patterns are truly discriminative for a specific instance, thereby limiting the reliability of instance-level interpretation.

Motivated by the above limitations, we propose \textit{INSHAPE} (\textit{IN}stance-level \textit{SHAPE}lets)\footnote{Code: \url{https://github.com/nshuhsn/INSHAPE_IJCAI}.}, a novel \textit{interpretable-by-design} framework for TSC. Instead of relying on a fixed set of globally optimized shapelets, INSHAPE discovers discriminative temporal patterns directly from each input time series while capturing their temporal dependencies.   
Specifically, INSHAPE first partitions each time series into variable-length, statistically coherent time-series segments. 
It then employs a gating mechanism -- comprising an amortized selector and a temporal predictor -- to identify segments that are most predictive of the target label. This enables the discovery of \textit{instance-level shapelets} that account for complex temporal interactions, providing local explanations that faithfully reflect the input's temporal structure as shown in Figure~\ref{fig1:instance_vs_population}(b).
Extensive experiments on 128 UCR and 30 UEA benchmark datasets demonstrate that INSHAPE achieves superior classification performance compared to state-of-the-art shapelet-based models. 
Furthermore, INSHAPE offers a novel pathway to global interpretability. By aggregating discovered discriminative instance-level shapelets across the dataset, our framework constructs a set of high-quality, population-level shapelets. 
When combined with our selector's localization capability, these population-level shapelets reveal not only which global patterns are identified, but also where and how they manifest across instances, bridging the gap between local and global interpretability.

\section{Related Work}\label{sec:related_work}

\subsection{Population-level Shapelet Discovery}
In TSC, shapelets~\cite{koegh_Shapelet_KDD2009} are widely used to provide global interpretability by identifying temporal patterns that frequently occur in specific classes. 
Early shapelet discovery methods relied on exhaustive search to evaluate all possible candidate subsequences, which is computationally prohibitive for large datasets. 
To address this, more recent works have focused on improving efficiency by pruning candidates with symbolic aggregate approximation~\cite{lin_SAX_2007,rakthanmanon_fastshapelet_SIAM2013}, clustering candidate patterns~\cite{grabocka_scalable_2015,zuo2_SVPT_AAAI2023,li_shapenet_AAAI2021}, and learning shapelets directly as learnable parameters via gradient descent~\cite{grabocka_LTS_KDD2014,qu_ShapeConv_ICLR2024,wen_InterpGN_ICLR2025}. 
Despite these advances, existing methods treat candidate segments independently during the discovery process. Hence, by ignoring the temporal dependencies among segments, these methods are restricted to interpreting individual patterns in isolation, failing to capture how the collective interaction and ordering of multiple patterns jointly contribute to the final classification.

\subsection{Population-level Shapelets for Classification}
Shapelet transform-based methods~\cite{hills_ShapeletTransform_DMKD2014,lines_ShapeletTransform2_KDD2012} leverage population-level shapelets for instance-level prediction and interpretation by representing each time series as a vector of minimum distances from a fixed set of shapelets. The resulting distance-based representation of each time series is then utilized by various types of classifiers, including linear models~\cite{grabocka_LTS_KDD2014,wen_InterpGN_ICLR2025}, 
SVMs~\cite{li_shapenet_AAAI2021}, and neural networks~\cite{qu_ShapeConv_ICLR2024,zuo2_SVPT_AAAI2023}, which learn class-discriminative decision boundaries based on the relative similarity.
Local interpretability is typically provided through post-hoc analysis by overlaying each shapelet at its closest matching location within a given time-series instance. 
However, such post-hoc overlays often fail to capture instance-specific discriminative patterns, leading to local explanations that are weakly aligned with the true temporal patterns of individual time series.
Moreover, because the shapelet transform primarily encodes the \textit{presence} of shapelets via distance-based representations, positional information and temporal dependencies among multiple shapelets are often discarded~\cite{mohammadi_deepTSCsurvey2024} (See Supplementary~A).
While recent work attempts to mitigate this limitation by incorporating population-level shapelets into deep neural networks that model temporal dependencies for TSC~\cite{zuo2_SVPT_AAAI2023}, these dependencies are still not considered during the shapelet discovery stage.

Unlike prior works, INSHAPE selects discriminative segments \textit{at the instance level} while accounting for their temporal dependencies to predict the target class during the selection process.
The most closely related work is SoftShape~\cite{liu_SoftShape_ICML2025}, which partitions time-series segments into top-$k$ ``important'' and ``unimportant'' groups, encoding them separately and modeling their interactions to construct the final prediction. 
However, because unimportant segments -- which do not correspond to any shapelet patterns -- still contribute to the prediction, it remains unclear which specific temporal patterns are truly responsible for the TSC. 
In contrast, INSHAPE employs \textit{hard gating} via Bernoulli sampling~\cite{maddison_concrete_2016,yoon_invase_2018,SynFS_Kim_ICML,SEFS_LEE_ICLR}. This stochastic selection process learns to flexibly select any number of important time-series segments while ensuring the classification depends exclusively on the selected segments for each time-series instance. Consequently, INSHAPE captures complex temporal interactions among instance-level shapelets whose contributions are directly attributable, thereby enabling clearer and more reliable local interpretations.

\section{Problem Formulation}
Consider a dataset comprising $n$ time-series instances, 
i.e., $\data= \big\{ (\mathbf{x}_{1:T}^{(i)}, y^{(i)}) \big \}_{i=1}^{n}$ 
where $\mathbf{x}_{1:T} = (x_1, \ldots, x_T) \in \mathbb{R}^{T}$ denotes 
a univariate time series of length $T$ and $y \in \{1, \ldots, C\}$ is 
its associated label in a $C$-class classification task. 
For notational simplicity, we focus on the univariate setting; however, our formulation naturally extends to $p$-dimensional multivariate time series $\mathbf{x}_{1:T}^{(i)} \in \mathbb{R}^{p \times T}$. 
We will omit the dependency for the instance index $(i)$ when it is clear from the context. 
\\
\begin{definition}[Partial Segmentation]
Given a time-series instance $\mathbf{x}_{1:T}$, a valid \textit{partial segmentation} divides it into $M$ non-overlapping segments, each potentially of varying length, as described below:
\begin{equation} \label{eq:partial_segmentation}
    \mathbf{s}(\mathbf{x}_{1:T}) \ldefeq (s_{1}, \dots, s_{M}),
\end{equation}
where each segment $s_m = (\mathbf{x}_{\tau_{m}^{s}}, \dots, \mathbf{x}_{\tau_{m}^{e}})$ with $\tau_{m}^{s} \leq \tau_{m}^{e} < \tau_{m+1}^{s}$ for all $m \in \{1, \dots, M-1\}$. The number of segments $M$ can range from $1$ to $T$, i.e., $1 \leq M \leq T$.
\end{definition}

A time series is considered \textit{fully segmented} if $\tau_{m+1}^{s} = \tau_{m}^{e} + 1$ for all $m \in \{1, \dots, M-1\}$, and we denote such a segmentation as $\bar{\sv}(\xv_{1:T}) =  (\bar{s}_{1}, \dots, \bar{s}_{M})$.
Unlike full segmentation, partial segmentation selects only a subset of the time series, such that the total number of selected time points may be less than $T$, i.e., $\sum_{m=1}^{M} |s_m| \leq T$. This reflects our focus on identifying only the most informative regions rather than covering the entire sequence.
Each time-series instance can have multiple valid partial segmentations that vary not only in the number of segments but also in their individual lengths. Denote $\mathcal{S}(\mathbf{x}_{1:T})$ be the set of all such valid partial segmentations of $\mathbf{x}_{1:T}$. Then, we can formally define instance-level shapelets as follows:

\begin{definition}[Instance-level Shapelets]
The instance-level shapelet of a time series $\mathbf{x}_{1:T}$ is defined as the most discriminative partial segmentation with respect to the target label:
\begin{equation} \label{eq:instance_level_shapelets}
   \mathbf{s}^{*} = (s_{1}^{*}, \dots, s_{M^*}^* ) = \underset{\sv \in \Sc(\xv_{1:T})}{\mathrm{argmax}}~ p(y \mid \mathbf{s})
\end{equation}
subject to the constraints $\sum_{m=1}^{M^*} | s_{m}^* | \leq \delta_1$ and $|s_{m}^{*} | \geq \delta_2$ for $m \in \{1,\dots, M^* \}$, where $\delta_1$ limits the total length of selected segments and $\delta_2$ constrains a minimum segment length. These constraints ensure that the selected segments remain concise and provide compact interpretations.
\end{definition}

\noindent \textbf{Challenges.}~
Unfortunately, solving the optimization problem in \eqref{eq:instance_level_shapelets} is highly non-trivial for several reasons. First, it requires estimating the objective function -- specifically, the conditional distribution of the target label given any arbitrary partial segmentation. Second, the space of possible partial segmentations for each instance, i.e., $\Sc(\xv_{1:T})$, is combinatorially large, making the discrete optimization computationally intractable. Third, the solution must satisfy interpretability constraints such that selecting a small number of segments while ensuring each segment is sufficiently long to capture semantically meaningful patterns.

\section{INSHAPE: INstance-level SHAPElet}
To overcome the challenges outlined above, we reformulate instance-level shapelet discovery as a problem of \textit{selecting discriminative segments} from a fully segmented time series. To do so, we introduce a gating mechanism designed to distinguish between discriminative and non-discriminative segments. Given a fully-segmented time series, $\bar{\sv}(\xv_{1:T})$, we define a binary \textit{gate vector} as follows:
\begin{equation}
    \gv = (g_1,\ldots, g_M)\in \{0,1\}^T,
\end{equation}
where each gate $g_m$ is a binary vector of length $\tau_m^e - \tau_m^s + 1$, set to all ones if the corresponding segment is selected, and all zeros otherwise. The instance-level shapelets of $\xv_{1:T}$ can then be discovered by optimizing the gate vector to maximize the following objective:
\begin{equation} \label{eq:instance_level_shapelets_gates}
   \gv^* = \underset{\gv \in \{0,1\}^T}{\mathrm{argmax}}~ p(y \mid \bar{\sv}(\xv_{1:T}) \odot \gv),
\end{equation}
where $\odot$ denotes element-wise multiplication. The resulting instance-level shapelet is thus given by $\sv^* = \bar{\sv}(\xv_{1:T}) \odot \gv^*$.

Solving this optimization problem in \eqref{eq:instance_level_shapelets_gates} presents two key challenges:
(\textbf{Time Series Segmentation}) We must fully segment each time-series instance into meaningful temporal patterns. This is crucial to avoid the computationally intractable burden of discrete optimization across all possible full segmentations. (\textbf{Amortized Gate Vector Optimization}) To manage the cost of solving a per-instance optimization problem, we amortize this expense across a population of input time series. This is achieved by training a neural network to learn input-dependent gate vectors for segment selection. 
We will describe each in turn in the following subsections. 

\begin{figure}[!t]
    \centering
    \includegraphics[width=\linewidth]{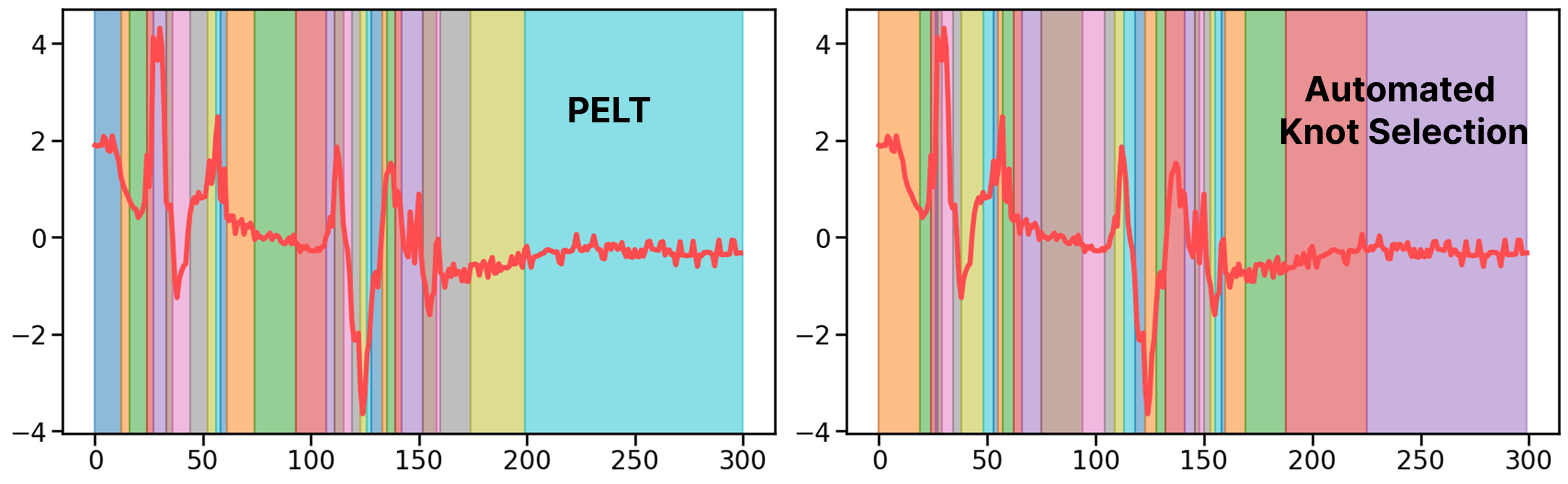}
    \caption{\textbf{Representative transition point algorithms.}
    Note that both algorithms consistently divide a given time series into similar statistically coherent regions.}
    \label{fig:algorithm}
\end{figure}

\subsection{Segmenting Time Series into Meaningful Patterns}\label{subsec:segmentation}
To reduce the combinatorial burden of exploring all possible full segmentations of a time-series instance, we instead focus on identifying meaningful transition points that delineate segment boundaries. Motivated by recent studies \cite{kacprzyk_TowardsTransparent_ICLR2024} that break down time series into interpretable patterns, we propose the following two properties that such transition points should satisfy for a desirable full segmentation: 
\begin{itemize}[leftmargin=*, topsep=0pt, partopsep=0pt, itemsep=0pt]
\item Transition points should be frequent in regions exhibiting abrupt and large fluctuations, as these often correspond to \textit{extreme values} informative for predicting the target class $y$. 
\item Transition points should be sparse in regions showing gradual, smooth variations, as such segments typically represent coherent \textit{trends} relevant to $y$. 
\end{itemize}
Hence, we adopt two representative transition point algorithms that reflect these properties from different perspectives: 

\noindent
\textbf{Automated Knot Selection \cite{dierckx_automatedKnots_SIAM1982}:} 
Denote $\Tc=\{\tau_1,\ldots,\tau_K\}$ be a set of knots (i.e., transition points) that partition $[0, T]$ into $K-1$ intervals, with B-spline basis functions $k_m(t)$ used on each interval $[\tau_m,\tau_{m+1}]$ for $m\in\{1, \dots, K-1\}$. To automatically determine the placement and number of these knots, we employ an iterative method that begins with a single segment (i.e., $K=2$). In each step, a new knot is inserted into the interval with the largest residual until the following inequality holds:
\begin{equation}\label{eq:knot}
\sum_{m=1}^{K-1}\sum_{t=\tau_m}^{\tau_{m+1}} c_1\!\left(k_m(t),x_t\right)\le \beta 
\end{equation}
 where $c_1(\cdot,\cdot)$ is the residual (e.g., squared error) and $\beta > 0$ is the smoothing parameter. 

\begin{figure*}[!t]
    \centering
    \includegraphics[width=0.95\textwidth]{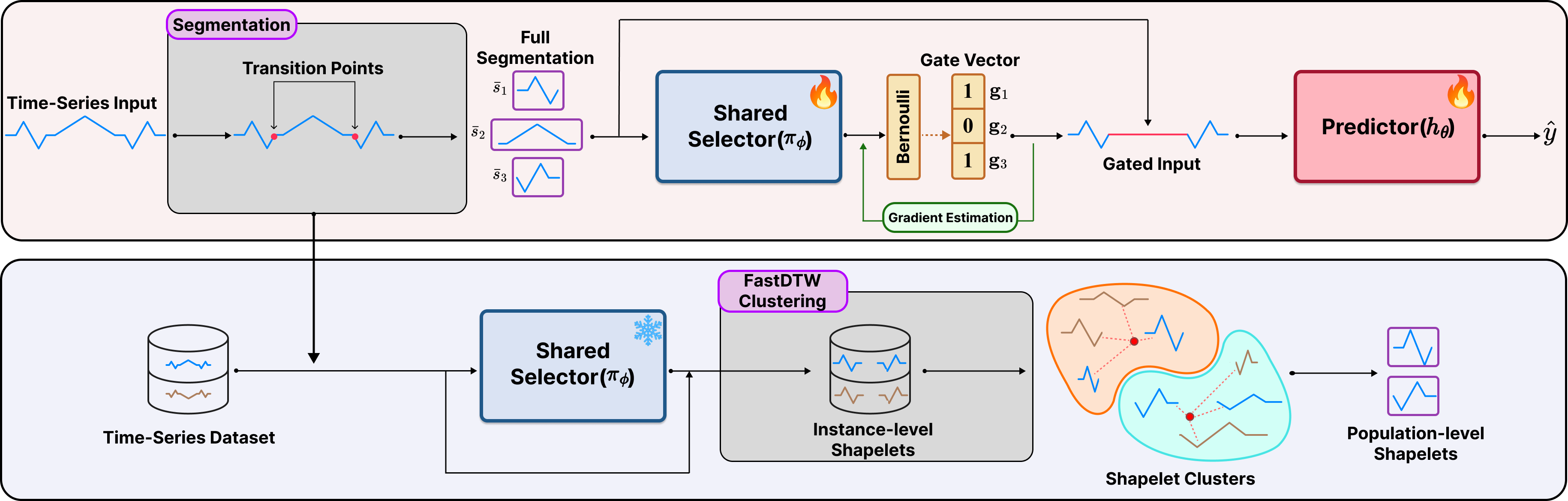}
    \caption{\textbf{Overview of our INSHAPE framework.} 
    \textbf{(Top)} During training, the transition point algorithm segments the input time series, and the shared stochastic selector $\pi_\phi$ learns a Bernoulli parameter for each segment to identify discriminative regions. 
    The gate vector $\mathbf{g}\sim\text{Bern}(\pi_\phi(\bar{\sv}))$ masks non-selected segments with zeros (while preserving positional information), and the predictor $h_\theta$ performs TSC based on the masked time series.
    To make gradient flow through the non-differentiable discrete Bernoulli sampling process, we employ a gradient estimator (i.e., ReinMax), allowing end-to-end training. 
    \textbf{(Bottom)} For population-level shapelet discovery, the pre-trained (frozen) selector extracts instance-level shapelets across the dataset, which are then clustered using FastDTW to obtain representative cluster centroids.} 
    \label{fig:framework_overview}
\end{figure*}
\noindent
\textbf{PELT Algorithm~\cite{killick_PELT_JASA2012}:}
The Pruned Exact Linear Time (PELT) algorithm identifies transition points 
by minimizing a penalized sum of segment-wise costs using dynamic programming with pruning:
\begin{equation}\label{eq:pelt}
\min_{\Tc}\;\sum_{m=1}^{K-1} c_2(\xv_{\tau_m:\tau_{m+1}})+\beta|\Tc|
\end{equation}
where $\Tc=\{\tau_1\ldots\tau_K\}$ is the set of transition points. The term $c_2(\cdot)$ represents the cost associated with each segment (e.g., using an RBF-based cost function), and $\beta>0$ penalizes the number of change points $|\Tc|$ to prevent over-segmentation.

Consequently, these transition point algorithms identify statistically coherent segments, $\bar\sv(\xv_{1:T})$, thereby avoiding the computational burden of full segmentation and providing segments with meaningful temporal patterns that serve as a basis for intuitive, reliable explanations~(Figure~\ref{fig:algorithm}).
In this work, we adopt PELT as our default segmentation algorithm due to its computational efficiency; the effect of utilizing B-spline and uniform segmentation is also provided in Section~\ref{Experiment_5_3}.

\subsection{Amortized Gate Vector Optimization}
The optimization in \eqref{eq:instance_level_shapelets_gates} is inherently \textit{memoryless}, requiring independent computation for each input time series without leveraging patterns from previously seen (and possibly similar) instances. To overcome this inefficiency, we reformulate the problem using amortized optimization by introducing a gating mechanism. Specifically, we define a selector (parameterized by $\phi$) that stochastically maps a varying length time-series segment $\bar{s}_m$ to a single binary gate $g_m$. 
To achieve the gate vector $\gv$, each segment is processed by the shared selector using a masked self-attention mechanism that captures the representation of its shape as illustrated in Figure~\ref{fig:framework_overview}.
Formally, the selector, $\pi_{\phi}: \mathbb{R}^{\tau_{m}^{e} -\tau_{m}^{s}+1} \rightarrow [0,1]$,  outputs a Bernoulli parameter based on $\bar{s}_m$, such that $g_{m} \sim \texttt{Bern}(\pi_\phi(\bar{s}_m))$. 
We also define a predictor $h_\theta: \mathbb{R}^T \rightarrow [0,1]^{C}$ that estimates the conditional distribution over class labels given any selected subset of segments.
This leads to the following joint optimization:
\begin{equation}\label{eq:final_objective}
\underset{\theta,\phi}{\mathrm{minimize}}~~\mathbb{E}_{\mathbf{x}_{1:T},y}~ \mathbb{E}_{\mathbf{g}} \Big[ \mathcal{L}\big(y, h_\theta(\bar{\sv}(\xv_{1:T})\odot\gv)\big) + \lambda \| \gv \|_0 \Big],
\end{equation}
where $\lambda > 0$ is a coefficient that balances the selected number of segments. 
The element-wise product $\bar{\sv}(\xv_{1:T}) \odot \gv$ masks non-selected segments with \textit{redundant zero-terms} rather than removing them, thereby preserving the original temporal context and positional information for the predictor. Additionally, since the discrete sampling of $\gv$ makes \eqref{eq:final_objective} non-differentiable with respect to $\phi$, we adopt ReinMax \cite{liu_REINMAX_NIPS2023}, a second-order accurate gradient estimator for binary random variables that reduces gradient bias compared to Straight-Through Estimators (STE)~\cite{bengio_STE_2013}. Notably, other gradient estimation techniques such as STGS \cite{jang_gumbel_2016}, REINFORCE \cite{williams_REINFORCE_1992} can also be applied within our framework.

We implement the selector, $\pi_\phi$, using Transformer~\cite{vaswani_transformer_NIPS2017} followed by an MLP layer, with both components shared across all segments to efficiently process variable-length segments in a single forward pass. For the predictor, $h_{\theta}$, we utilize an InceptionTime architecture~\cite{ismail_inceptiontime_DMKD2020}, whose multi-level convolutional filters are adept at capturing diverse dependencies among the selected segments; we further evaluate alternative predictor architectures in Section~\ref{Experiment_5_3}. 
Through joint optimization, the InceptionTime predictor generates rich and informative gradient signals that are propagated back to the selector. This feedback loop is crucial: it enables the selector to implicitly learn and account for inter-segment dependencies, even though the selector processes each segment individually. 

Once trained, INSHAPE provides \textit{local} interpretability for an individual time-series instance by first applying the segmentation algorithm and then leveraging the selector’s output; the segments corresponding to the activated gates are identified as instance-level shapelets.

\subsection{Bridging Local and Global Interpretability} \label{subsec: population-level Shapelets}
Local interpretability alone is not sufficient to answer population-level questions, such as identifying which temporal patterns are \textit{commonly important} for distinguishing between classes across the dataset. To address this, we adopt a \textit{select-then-cluster} strategy to derive \emph{population-level shapelets}. Specifically, we first use the trained selector to extract instance-level shapelets from all time-series instances. 
These selected segments are then clustered via FastDTW~\cite{salvador_FastDTW_IntelligentDataAnalysis2007}, which naturally handles varying-length time-series segments, and the resulting cluster centroids are defined as the population-level shapelets (see Figure~\ref{fig:framework_overview}).
By restricting the candidate pool to selected discriminative segments rather than clustering raw subsequences, INSHAPE substantially reduces the search space and enables the discovery of more representative and meaningful population-level shapelets.

Building on the discovered population-level shapelets, our framework goes beyond identifying global patterns by explicitly \textit{bridging local and global interpretations} and providing \textit{quantitative insights} into their class-wise correspondences:
\begin{itemize}[leftmargin=*, topsep=0pt, partopsep=0pt, itemsep=0pt] 
    \item \textbf{Bridging Local and Global Interpretations:} Each instance-level shapelet is assigned to its nearest population-level shapelet based on the FastDTW distance. This assignment connects global trends to local explanations by revealing where the discovered population-level shapelets manifest within individual instances (see Figure~\ref{fig:UCR_qualitative_figure}(b)).  
    \item \textbf{Quantitative Global Insights:} We compute the class-wise frequency of each population-level shapelet assigned to time-series instances. This reveals which global patterns commonly characterize each class. Visualizing these frequent patterns across the population provides insights into shared temporal structures and dependencies (see Figure \ref{fig:global_interpretation}).  
\end{itemize}
Detailed procedures for population-level shapelet discovery, local interpretation using population-level shapelets, and the computation of quantitative global insights are provided in Supplementary Material Section~B.4.
\newcommand{\HH}[1]{\textbf{\shortstack[c]{#1}}}
\newcommand{\MB}[1]{\mbox{#1}}

\begin{table}[!t]
\centering
\resizebox{\columnwidth}{!}{
\begin{tabular}{cccccc>{\columncolor[HTML]{FFF3F3}}c}
\toprule
& LTS & ShapeNet & ShapeConv & SVP-T & SBM & \textbf{Ours} \\
\midrule
IS & \xmark & \xmark & \xmark & \xmark & \xmark & \color{green}\cmark \\
PS & \color{green}\cmark & \color{green}\cmark & \color{green}\cmark & \color{green}\cmark & \color{green}\cmark & \color{green}\cmark \\
\makecell[c]{SD w/ TD} & \xmark & \xmark & \xmark & \xmark & \xmark & \color{green}\cmark \\
\makecell[c]{Pred w/ TD} & \xmark & \xmark & \xmark & \color{green}\cmark & \xmark & \color{green}\cmark \\
\bottomrule
\end{tabular}}
\caption{\textbf{Comparison of shapelet-based methods.} (IS: Instance-level shapelet, PS: Population-level shapelet, SD: Shapelet Discovery, Pred: Prediction, TD: Temporal Dependency)}
\label{tab:method_comparison}
\end{table}

\begin{table*}[!t]
\setcounter{table}{1}
\centering
\scriptsize
\setlength{\tabcolsep}{4pt}
\renewcommand{\arraystretch}{1.15}

\begin{minipage}[b]{0.56\textwidth}
\centering
\renewcommand{\arraystretch}{1.05}
\begin{adjustbox}{width=\linewidth}
\begin{tabular}{
  >{\raggedright\arraybackslash}m{1.4cm}
  *{6}{>{\centering\arraybackslash}m{0.95cm}}
  | >{\centering\arraybackslash\columncolor[HTML]{FFF3F3}}m{0.95cm}
}
\toprule
\multicolumn{8}{c}{\textbf{Univariate (UCR 128)}} \\
\midrule
\textbf{Metric}
& \textbf{LTS}
& \HH{Shape\\Net}
& \HH{SVP-T}
& \HH{Shape\\Conv}
& \textbf{SBM}
& \HH{Soft\\Shape}
& \textbf{Ours} \\
\midrule
Avg. Acc \color{red}$\uparrow$
& 0.6262 & 0.7213 & 0.7018 & 0.7479 & 0.7375 & \underline{0.7820} & \textbf{0.8405} \\
Avg. Rank \color{blue}$\downarrow$
& 5.797 & 4.371 & 4.277 & 4.117 & 3.938 & \underline{3.105} & \textbf{2.395} \\
\hline
Wins/Draws & 116 & 93 & 102 & 108 & 98 & 82 & -- \\
Losses     & 12  & 35 & 26  & 20  & 30 & 46 & -- \\
Top-1      & 2   & 30 & 16  & 9   & 29 & 0  & \textbf{46} \\
Top-3      & 11  & 46 & 48  & 50  & 49 & \underline{84} & \textbf{105} \\
\bottomrule
\end{tabular}
\end{adjustbox}

\begin{adjustbox}{width=\linewidth}
\begin{tabular}{
  >{\raggedright\arraybackslash}m{1.55cm}
  *{5}{>{\centering\arraybackslash}m{1.20cm}}
  | >{\centering\arraybackslash\columncolor[HTML]{FFF3F3}}m{1.20cm}
}
\toprule
\multicolumn{7}{c}{\textbf{Multivariate (UEA 30)}} \\
\midrule
\textbf{Metric}
& \textbf{LTS}
& \HH{Shape\\Net}
& \HH{SVP-T}
& \HH{Shape\\Conv}
& \textbf{SBM}
& \textbf{Ours} \\
\midrule
Avg. Acc \color{red}$\uparrow$
& 0.7150 & 0.6306 & 0.6783 & 0.6655 & \underline{0.7277} & \textbf{0.7429} \\
Avg. Rank \color{blue}$\downarrow$
& 3.183 & 4.950 & 3.700 & 4.050 & \underline{2.567} & \textbf{2.550} \\
\hline
Wins/Draws & 16 & 24 & 25 & 22 & 16 & -- \\
Losses     & 14 & 6  & 5  & 8  & 14 & -- \\
Top-1      & 3  & 3  & 3  & 5  & 10 & \textbf{11} \\
Top-3      & 19 & 6  & 13 & 10 & 22 & \textbf{23} \\
\bottomrule
\end{tabular}
\end{adjustbox}
\subcaption{Classification performance (Average Acc and Rank).}
\label{tab:classification-performance}
\end{minipage}%
\hfill
\begin{minipage}[b]{0.42\textwidth}
\centering
\renewcommand{\arraystretch}{1.265}
\begin{minipage}[t]{0.49\linewidth}
\centering
\begin{adjustbox}{width=\linewidth}
\begin{tabular}{
  >{\raggedright\arraybackslash}m{1.9cm}
  *{3}{>{\centering\arraybackslash}m{0.62cm}}
}
\toprule
\multicolumn{4}{c}{\textbf{Baseline Methods (UCR 128)}} \\
\midrule
\textbf{Metric (Avg.)} & \textbf{LTS} & \HH{Shape\\Net} & \textbf{SBM} \\
\midrule
\MB{Coverage (Top-1)} & 0.3950 & 0.2840 & 0.4478 \\
\rowcolor[HTML]{F8EEFF}\MB{Coverage (Top-2)} & 0.5564 & 0.4554 & 0.5872 \\
\rowcolor[HTML]{F8EEFF}\MB{Coverage (Top-3)} & 0.6493 & 0.5827 & 0.6806 \\
\MB{Coverage (Full)}  & 0.9852 & 0.8935 & 0.9814 \\
\midrule
\MB{Overlap (Top-1)}  & 0.0000 & 0.0000 & 0.0000 \\
\rowcolor[HTML]{F8EEFF}\MB{Overlap (Top-2)}  & 0.2734 & 0.1310 & 0.3033 \\
\rowcolor[HTML]{F8EEFF}\MB{Overlap (Top-3)}  & 0.4160 & 0.2581 & 0.4398 \\
\MB{Overlap (Full)}   & 0.9727 & 0.7685 & 0.9679 \\
\midrule
\MB{Acc (Top-1)}      & 0.4163 & 0.3246 & 0.4016 \\
\rowcolor[HTML]{F8EEFF}\MB{Acc (Top-2)}      & 0.4089 & 0.3439 & 0.4147 \\
\rowcolor[HTML]{F8EEFF}\MB{Acc (Top-3)}      & 0.4025 & 0.3438 & 0.4217 \\
\MB{Acc (Full)}       & 0.6262 & 0.7213 & 0.7375 \\
\bottomrule
\toprule
\rowcolor[HTML]{FFF3F3}\multicolumn{4}{c}{\textbf{Ours (UCR 128)}} \\
\midrule
\rowcolor[HTML]{FFF3F3}Avg. Shapelet Num & \multicolumn{3}{c}{\textbf{2.4151}} \\
\rowcolor[HTML]{FFF3F3}Avg. Coverage     & \multicolumn{3}{c}{\textbf{0.4949}} \\
\rowcolor[HTML]{FFF3F3}Avg. Overlap      & \multicolumn{3}{c}{\textbf{0.0000}} \\
\rowcolor[HTML]{FFF3F3}Avg. Acc          & \multicolumn{3}{c}{\textbf{0.8405}} \\
\bottomrule
\end{tabular}
\end{adjustbox}
\end{minipage}%
\hfill%
\begin{minipage}[t]{0.49\linewidth}
\centering
\begin{adjustbox}{width=\linewidth}
\begin{tabular}{
  >{\raggedright\arraybackslash}m{1.9cm}
  *{3}{>{\centering\arraybackslash}m{0.62cm}}
}
\toprule
\multicolumn{4}{c}{\textbf{Baseline Methods (UEA 30)}} \\
\midrule
\textbf{Metric (Avg.)} & \textbf{LTS} & \HH{Shape\\Net} & \textbf{SBM} \\
\midrule
\rowcolor[HTML]{F8EEFF}\MB{Coverage (Top-1)} & 0.4379 & 0.2953 & 0.4943 \\
\rowcolor[HTML]{F8EEFF}\MB{Coverage (Top-2)} & 0.5813 & 0.4546 & 0.6487 \\
\MB{Coverage (Top-3)} & 0.6769 & 0.5469 & 0.7290 \\
\MB{Coverage (Full)}  & 0.9845 & 0.7289 & 0.9797 \\
\midrule
\rowcolor[HTML]{F8EEFF}\MB{Overlap (Top-1)}  & 0.0000 & 0.0000 & 0.0000 \\
\rowcolor[HTML]{F8EEFF}\MB{Overlap (Top-2)}  & 0.2910 & 0.1301 & 0.3490 \\
\MB{Overlap (Top-3)}  & 0.4236 & 0.2432 & 0.4934 \\
\MB{Overlap (Full)}   & 0.9714 & 0.5760 & 0.9653 \\
\midrule
\rowcolor[HTML]{F8EEFF}\MB{Acc (Top-1)}      & 0.2345 & 0.3072 & 0.2516 \\
\rowcolor[HTML]{F8EEFF}\MB{Acc (Top-2)}      & 0.2496 & 0.2868 & 0.2784 \\
\MB{Acc (Top-3)}      & 0.2497 & 0.3253 & 0.2921 \\
\MB{Acc (Full)}       & 0.7150 & 0.6306 & 0.7277 \\
\bottomrule
\toprule
\rowcolor[HTML]{FFF3F3}\multicolumn{4}{c}{\textbf{Ours (UEA 30)}} \\
\midrule
\rowcolor[HTML]{FFF3F3}Avg. Shapelet Num & \multicolumn{3}{c}{\textbf{1.5266}} \\
\rowcolor[HTML]{FFF3F3}Avg. Coverage     & \multicolumn{3}{c}{\textbf{0.5539}} \\
\rowcolor[HTML]{FFF3F3}Avg. Overlap      & \multicolumn{3}{c}{\textbf{0.0000}} \\
\rowcolor[HTML]{FFF3F3}Avg. Acc          & \multicolumn{3}{c}{\textbf{0.7429}} \\
\bottomrule
\end{tabular}
\end{adjustbox}
\end{minipage}
\subcaption{Local interpretability (Coverage and Overlap).}
\label{tab:interpretability-metrics}
\end{minipage}

\caption{\textbf{Comparison of predictive performance and interpretability on the UCR (128 datasets) and UEA (30 datasets) benchmarks.}}
\label{tab:overall-stats-interpretability}
\end{table*}

\section{Experiments}
We evaluate INSHAPE on the UCR archive (128 univariate datasets)~\cite{dau_ucr_JAS2019} and the UEA archive (30 multivariate datasets)~\cite{bagnall_UEA_arxiv}, comparing against six representative shapelet-based methods summarized in Table~\ref{tab:method_comparison}.
All experiments follow established evaluation protocols from prior works~\cite{liu_SoftShape_ICML2025,wen_InterpGN_ICLR2025}.

\noindent 
\begin{itemize}[leftmargin=*, topsep=0pt, partopsep=0pt, itemsep=0pt] 
    \item In Section~\ref{Experiment_5_1}, we compare classification accuracy and interpretability against baselines, showing that INSHAPE achieves superior performance with clearer interpretations.  
    \item In Section~\ref{Experiment_5_2}, we show how population-level shapelets can be used for local and global interpretation. 
    \item In Section~\ref{Experiment_5_3}, we conduct ablation studies to analyze the effects of segmentation strategies and predictor architectures. 
\end{itemize}  
Implementation details, baseline configurations, and metric definitions are provided in Supplementary Material Section~B.

\subsection{Classification Performance and Interpretability}\label{Experiment_5_1}
\noindent
\textbf{Performance Analysis.}~
Table~\ref{tab:overall-stats-interpretability}(a) summarizes the average classification accuracy and ranking statistics on the UCR and UEA datasets. For a fair comparison, we use the interpretable variant of SoftShape; details of this modification and comparison with the original model are provided in Supplementary Material Section~B. Across all baselines, INSHAPE achieves the highest average accuracy and the best overall ranking. This highlights that preserving instance-specific information and modeling temporal dependencies are crucial for accurate TSC.

\begin{figure}[!t]
    \centering
    \includegraphics[width=0.90\linewidth, trim={10mm 11mm 0 0}, clip]{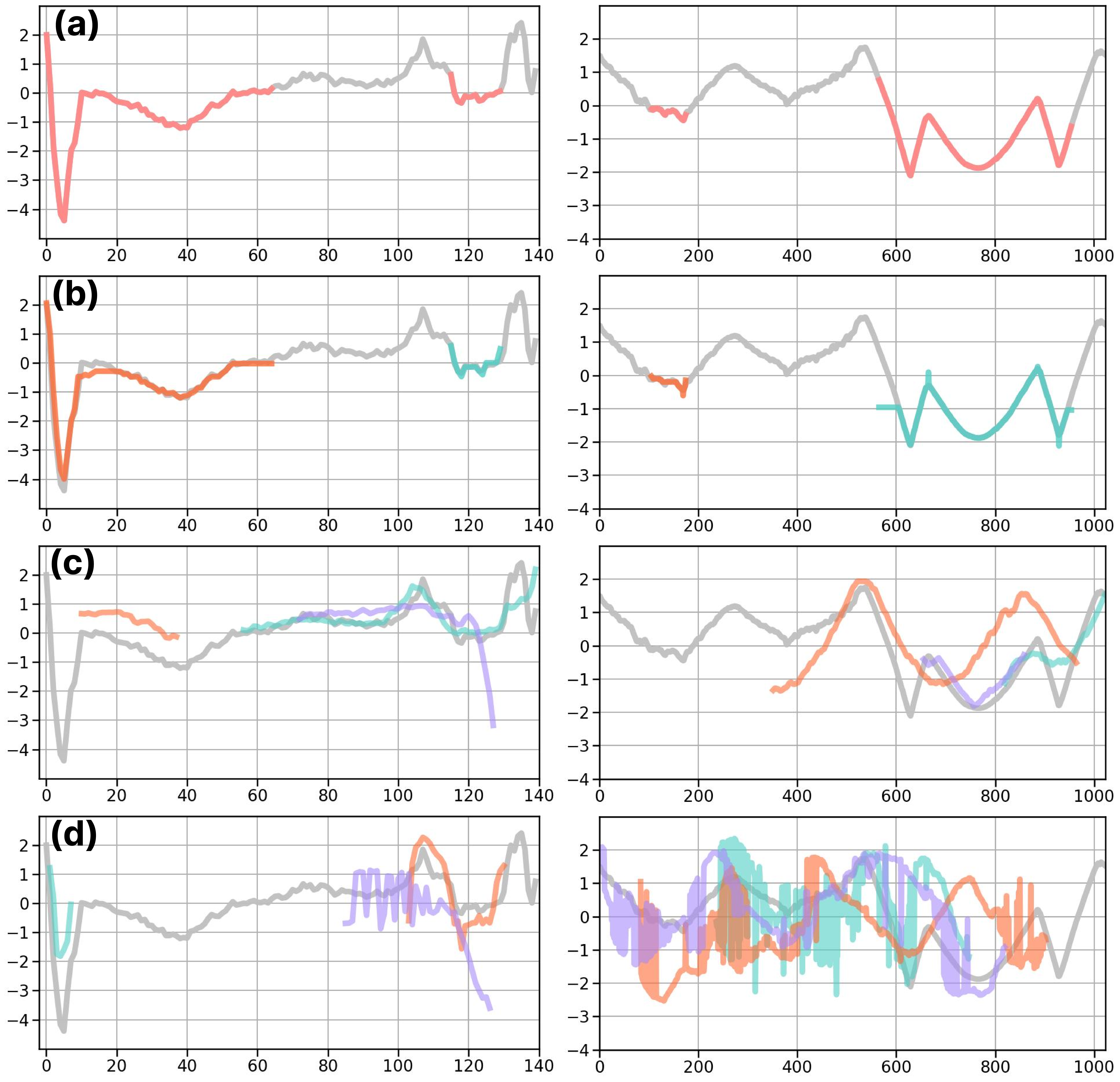} 
    \caption{\textbf{Local explanations on ECG5000 (Left) and MixedShapesRegularTrain (Right).} Selected shapelets from (a) INSHAPE, (b) INSHAPE (population), (c) ShapeNet, and (d) SBM are overlaid on the test instance. INSHAPE provides clearer local interpretations with non-overlapping segments.}
    \label{fig:UCR_qualitative_figure} 
\end{figure}

\begin{table}[!t]
\centering
\resizebox{\columnwidth}{!}{
\begin{tabular}{ccccccc|>{\columncolor[HTML]{FFF3F3}}c}
\toprule
& LTS & \begin{tabular}{@{}c@{}}Shape\\Net\end{tabular} & SVP-T & \begin{tabular}{@{}c@{}}Shape\\Conv\end{tabular} & SBM & \begin{tabular}{@{}c@{}}Soft\\Shape\end{tabular} & \textbf{\makecell{Ours \\ (Pop.)}}\\
\midrule
Avg. Acc \color{red}$\uparrow$& 0.6542 & 0.7601 & 0.7233 & 0.7876 & 0.772 & 0.7863 & \textbf{0.8145} \\
Avg. Rank \color{blue}$\downarrow$& 5.944 & 3.889 & 4.278 & 3.444 & 3.778 & 3.500 & \textbf{3.167} \\
\bottomrule
\end{tabular}
}
\caption{\textbf{Classification performance} using population-level shapelets in place of instance-level shapelets on 18 UCR datasets.}
\label{tab:I2P_ucr18}
\end{table}

\noindent
\textbf{Local Interpretability Analysis.}~
For local interpretation, existing shapelet-based methods typically overlay discovered shapelets on the input time series at positions with the smallest distance.
We compare local interpretability between INSHAPE and three representative shapelet transform-based baseline methods on the UCR and UEA datasets in Table~\ref{tab:overall-stats-interpretability}(b).
To quantify this, we measure the following metrics for the top-$k$ most important shapelets (ranked by absolute linear classifier weights) and the complete set of all available shapelets:
\begin{itemize}[leftmargin=*, topsep=0pt, partopsep=0pt, itemsep=0pt] 
    \item \textbf{Coverage}: Proportion of the input time series covered by shapelets. 
    Lower coverage indicates more compact explanations that focus on the most critical temporal regions. 
    \item \textbf{Overlap}: Proportion of temporal overlap among selected shapelets.
    Lower overlap provides clearer interpretations by avoiding redundant highlighting of the same regions. 
    \item \textbf{Shapelet Num}: Average number of instance-level shapelets selected per instance. 
\end{itemize}
In multivariate settings, Coverage and Overlap are measured separately for each channel and then averaged across channels.

In Table~\ref{tab:overall-stats-interpretability}(b), INSHAPE selects an average of $2.4$ and $1.5$ instance-level shapelets per time series on the UCR and UEA datasets, respectively. We therefore compare INSHAPE against baselines using their top-$1$, top-$2$ and top-$3$ shapelets. Under these sparse settings, INSHAPE achieves comparable coverage to baselines while exhibiting zero overlap by architectural design, clearly identifying distinct discriminative regions per instance. In contrast, baselines show substantial overlap (13-44\% on UCR, 0-35\% on UEA) even across the most important shapelets, obscuring which patterns drive predictions. 
Restricting baseline methods to make predictions solely on the top-$k$ shapelets leads to a substantial drop in classification accuracy, as evidenced by the top-1 and top-2 results. To recover the original performance, these methods need to utilize the entire shapelet set. This results in near-complete coverage ($>$98\%) and overlap ($>$96\%) for a given time-series instance, demonstrating that a sparse, focused local interpretation is not achievable with these methods.
In contrast, INSHAPE achieves superior prediction performance while relying exclusively on a small set of non-overlapping temporal regions for each time series, thereby providing clearer and more intuitive local interpretations.
Figure~\ref{fig:UCR_qualitative_figure} shows this distinction: baseline methods produce overlapping, misaligned explanations, while INSHAPE provides compact, non-overlapping segments that offer clearer explanations. Similar trends are observed in the UEA datasets (see Supplementary Material~Section C.1).

\subsection{Instance-level to Population-level Shapelets}
\label{Experiment_5_2}
In this subsection, we evaluate the population-level shapelets derived by INSHAPE. 
We derive population-level shapelets, denoted as $\mathcal{P} = \{p_1, \ldots, p_{n_{\mathcal{P}}}\}$, by clustering the discriminative instance-level shapelets extracted from all training time-series instances as described in Section \ref{subsec: population-level Shapelets} (see Supplementary Material Algorithm~1 for details).

\noindent
\textbf{Faithfulness of Population-level Shapelets.} To validate that population-level shapelets preserve instance-level discriminative information, we replace each instance-level shapelet with its nearest population-level counterpart (based on FastDTW distance), mask all non-selected regions, and then evaluate classification performance. As shown in Table~\ref{tab:I2P_ucr18}, INSHAPE with population-level shapelets achieves the highest average accuracy (0.8145) and the best average rank (3.167), confirming that our bottom-up approach produces faithful population-level shapelets.

\noindent
\textbf{Quantitative Global Insights.} To provide \textit{global insights}, we compute the usage frequency $u_{c,k}$ for each class $c$, by counting the number of instance-level shapelets from class-$c$ instances that are assigned to each population-level shapelet $p_k$.
The normalized frequency $\bar{u}_{c,k} = u_{c,k} / \sum_{k'} u_{c,k'}$ quantifies how frequently each class relies on specific population-level shapelets (see Supplementary Material Algorithm~3).

Figure~\ref{fig:global_interpretation} illustrates this global interpretation on the ECG5000 dataset.
We first select the top-3 most frequently used shapelets for each class and take their union, resulting in 5 representative shapelets: $\{p_0, p_1, p_5, p_8, p_{12}\}$. 
Figure~\ref{fig:global_interpretation}(a) presents the class-wise usage proportions $\bar{u}_{c,k}$. 
Although these population-level shapelets are shared across classes, their usage proportions differ substantially, indicating that class distinctions arise not only from \textit{which} shapelets appear, but also from their \textit{combinations} and \textit{relative frequencies}.

To further investigate these differences, Figure~\ref{fig:global_interpretation}(b) and (c) illustrate population-level shapelets overlaid on the time-series instances with Class~0 and 4, which have a similar shapelet composition but with different proportions.
While both classes are characterized by Shapelet~1 (pink) appearing in the early time steps, Class~0 predominantly relies on Shapelet~12 (light blue) in later time steps, whereas Class~4 relies more on Shapelet~5 (green).
This comparison underscores that different temporal dependencies among the same set of shapelets serve as a critical discriminative factor between classes.
Moreover, the localized regions identified for Class~0 closely align with the instance-level discriminative regions shown in Figure~\ref{fig:UCR_qualitative_figure}, demonstrating consistency between local and global interpretations.
Overall, these results show that our framework extends naturally beyond instance-level explanations, enabling rich global insights such as class-wise shapelet utilization and analysis of temporal dependencies.

\begin{figure}[!t]
\centering
\includegraphics[width=0.90\linewidth]{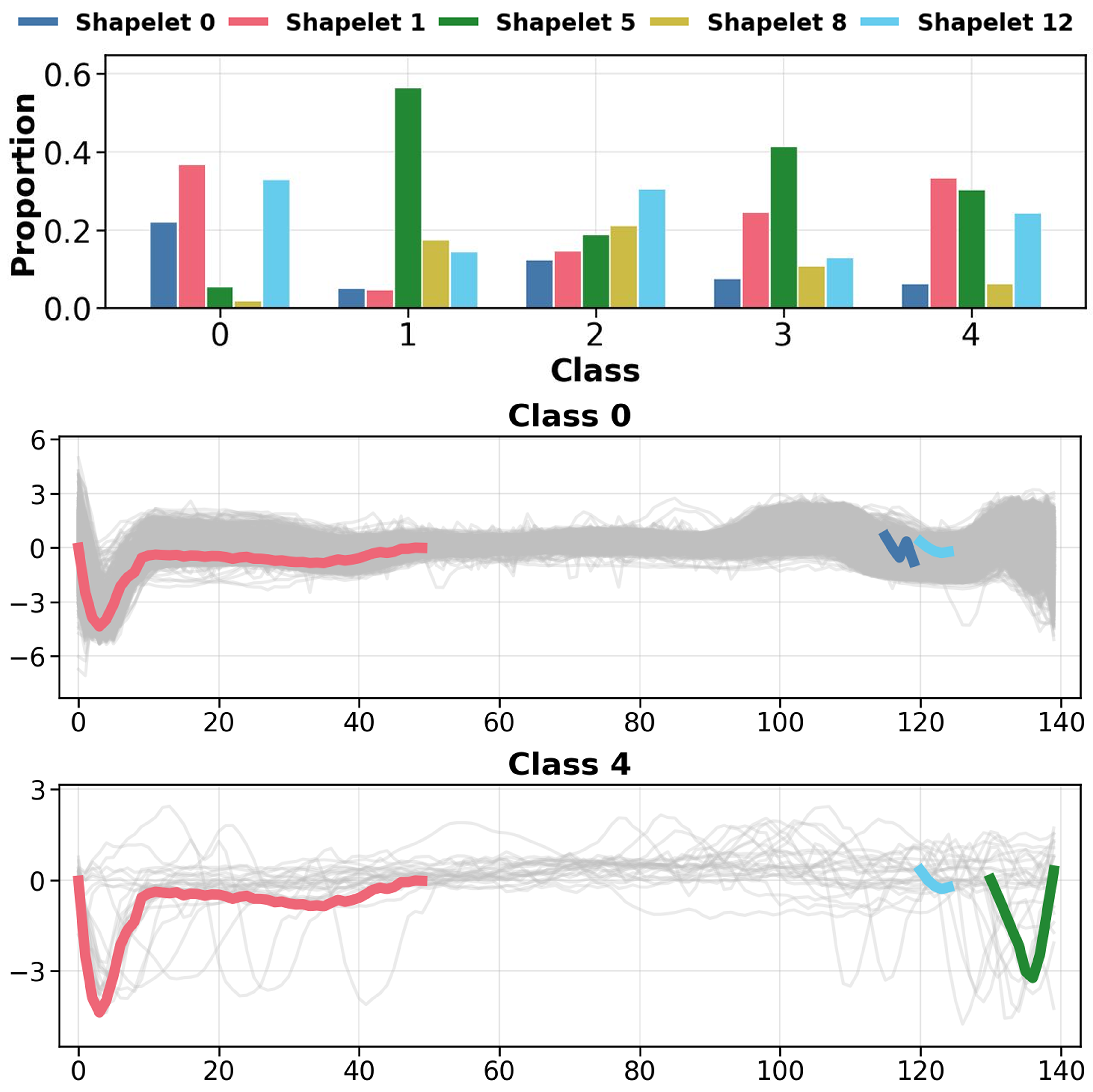} 
\caption{\textbf{Global interpretation via population-level shapelets on the ECG5000 dataset.} (a) Class-wise usage frequency $\bar{u}_{c,k}$ for 5 representative shapelets. (b)-(c) Visualization of population-level shapelets (colored) overlaid over time-series instances (gray) for Class~0 and Class~4.}
\label{fig:global_interpretation} 

\end{figure}

\subsection{Ablation Study}\label{Experiment_5_3}
We perform ablation studies on the penalty parameter $\beta$ in the PELT algorithm, the choice of transition point algorithms, and the predictor architectures, using the 18 UCR datasets following prior work~\cite{liu_SoftShape_ICML2025}.

\noindent\textbf{Sensitivity Analysis on the Segmentation Parameter.} We use PELT~\cite{killick_PELT_JASA2012} as the default transition point algorithm. To ensure consistent segmentation behavior across datasets with varying sequence lengths, we normalize the penalty parameter $\beta$ in \eqref{eq:pelt} as $\bar{\beta} = \beta / T$ and tune $\bar{\beta}$ as the hyperparameter. The actual penalty is then computed as $\beta = \bar{\beta} \times T$.
Table~\ref{tab:alpha_comparison} summarizes the average ACC, coverage, and the number of discovered shapelets under different $\bar{\beta}$ values.
When $\bar{\beta}$ is too small, the algorithm produces excessive transition points, leading to over-segmentation that fragments even smooth temporal trends into meaningless patterns. Conversely, when $\bar{\beta}$ is too large, the algorithm becomes overly conservative and misses important transitions.
The default ($\bar{\beta}=0.1$) provides the best trade-off, confirming that the two transition point properties for desirable segmentation described in Section~\ref{subsec:segmentation} are critical for the instance-level shapelet discovery.

\noindent\textbf{Segmentation algorithm comparison.} Table~\ref{tab:ablation_segmentation} compares fixed-length segmentation against adaptive methods (B-Spline and PELT). 
Adaptive methods outperform fixed-length approaches across all metrics, achieving higher accuracy with fewer shapelets. These results confirm that statistically coherent regions identified by adaptive segmentation align well with discriminative patterns in time series.

\noindent\textbf{Predictor architecture.} To demonstrate INSHAPE's robustness across different predictor architectures, we evaluate performance using two recent time-series classification models as the predictor~\cite{luo_moderntcn_ICLR2024,wu_timesnet_ICLR2023}.
As shown in Table~\ref{tab:predictor_comparison}, INSHAPE maintains strong performance with both architectures, demonstrating its compatibility with diverse predictors that can capture temporal dependencies to provide informative gradient signals to the selector.

\begin{table}[!t]
\centering
\footnotesize
\setlength{\tabcolsep}{4pt}
\renewcommand{\arraystretch}{1.0}
\resizebox{0.72\columnwidth}{!}{%
\begin{tabular}{cccc}
\toprule
$\bar{\beta}$ & Avg. Acc \color{red}$\uparrow$& Avg. Coverage & Avg. Shapelet Num \\
\midrule
0.01 & 0.8370 & 0.5347 & 2.3371 \\
~\textbf{0.1}$^*$ & 0.8677 & 0.4834 & 2.7396 \\
1 & 0.8230 & 0.5591 & 1.7816 \\
5 & 0.8444 & 0.6979 & 1.1171 \\
10 & 0.7901 & 0.8170 & 0.9358 \\
\bottomrule
\end{tabular}%
}
\caption{\textbf{Ablation study on the (normalized) penalty parameter $\bar{\beta}$.}}
\label{tab:alpha_comparison}
\end{table}

\begin{table}[!t]
\centering
\footnotesize
\setlength{\tabcolsep}{4pt}
\renewcommand{\arraystretch}{1.05}
\resizebox{0.72\columnwidth}{!}{%
\begin{tabular}{lcccc}
\toprule
 & \multicolumn{2}{c}{Fixed Length} & \multicolumn{2}{c}{Segmentation} \\
\cmidrule(lr){2-3} \cmidrule(lr){4-5}
Metric & W = 1 & W = 2 & B-Spline & PELT \\
\midrule
Avg. Acc \color{red}$\uparrow$ & 0.7992 & 0.7666 & \textbf{0.8288} & \textbf{0.8677} \\
Avg. Rank \color{blue}$\downarrow$ & 2.6389 & 3.1944 & \textbf{2.2222} & \textbf{1.9444} \\
Avg. Coverage & 0.5299 & 0.5146 & \textbf{0.5019} & \textbf{0.4834} \\
Avg. Shapelet Num & 5.4933 & 4.4152 & \textbf{2.6570} & \textbf{2.7396} \\
\bottomrule
\end{tabular}%
}
\caption{\textbf{Ablation study on transition point algorithms.}}
\label{tab:ablation_segmentation}
\end{table}

\begin{table}[!t]
\centering
\footnotesize
\setlength{\tabcolsep}{4pt}
\renewcommand{\arraystretch}{1.0}
\resizebox{0.60\columnwidth}{!}{%
\begin{tabular}{lcc}
\toprule
 & TimesNet & ModernTCN \\
\midrule
Avg. Acc \color{red}$\uparrow$ & 0.7855 & 0.8209 \\
Avg. Coverage & 0.5569 & 0.5295 \\
Avg. \#S & 2.4431 & 2.3396 \\
\bottomrule
\end{tabular}%
}
\caption{\textbf{Ablation study on predictor architectures.}}
\label{tab:predictor_comparison}
\end{table}

\section{Conclusion} 
We propose INSHAPE, an interpretable-by-design framework for TSC that discovers instance-level shapelets through variable-length segmentation and a gating mechanism, enabling precise alignment with input time series and capturing the underlying temporal dependencies. Furthermore, by adopting a select-then-cluster strategy, INSHAPE derives population-level shapelets that provide better global insight. This unified perspective is particularly valuable for high-stakes domains such as healthcare. 

\newpage

\section*{Ethical Statement}
While our framework offers both local and global interpretability for TSC through both instance-level and population-level shapelets, the discovered temporal patterns should be validated by domain experts before deployment in high-stakes applications. This is particularly important in healthcare settings, where our method could be used to identify temporal biomarkers or support clinical decision-making. We emphasize that INSHAPE is intended to augment, not replace, expert judgment. In this work, we evaluate our method on publicly available UCR and UEA benchmark datasets, which do not contain personally identifiable information. Any future application to sensitive domains such as clinical diagnostics should follow appropriate ethical guidelines and data governance protocols established by the respective institutions.



\section*{Acknowledgements}
We thank the reviewers for their comments and suggestions.
We also thank Professor Eamonn Keogh of the University of California, Riverside (UCR), and the contributors to the UCR Time Series Classification Archive.\\
\noindent
This work was supported by the National Research Foundation of Korea (NRF) grant funded by the Korea government (MSIT) (No. RS-2024-00358602) and the Institute of Information \& Communications Technology Planning \& Evaluation (IITP) grant funded by the Korea government (MSIT), Artificial Intelligence Graduate School Program (No. RS-2019-II190079, Korea University), the Artificial Intelligence Star Fellowship Support Program to nurture the best talents (No. RS-2025-02304828), and the AI Research Hub Project (No. RS-2024-00457882).






\bibliographystyle{named}
\bibliography{main}

\newpage
\clearpage

\appendix
\section{Synthetic: Latent Problem of Modern Shapelet Transform Approaches}
\paragraph{Review of Shapelet Transform.}
Shapelets are discriminative subsequences that are representative of a class in time-series classification.
The shapelet transform (ST) method~\cite{hills_ShapeletTransform_DMKD2014,lines_ShapeletTransform2_KDD2012} converts a raw time-series input $\xv\in\R^T$ into an interpretable feature representation by measuring its similarity to a set of shapelets $\sv = \{s_1,\ldots, s_n\}$.
Specifically, for each shapelet $s_k$, the transform computes a distance feature $d_k$ as:
\begin{equation}
    d_k = \min_{i} \text{dist}(\xv_{i:i+|s_k|},s_k)
\end{equation}
where $\xv_{i:i+|s_k|}$ denotes a subsequence of $\xv$ starting at position $i$ with length $|s_k|$, 
and $\text{dist}(\cdot,\cdot)$ is typically the Euclidean distance. The resulting distance vector $\mathbf{d}=\{d_1,\ldots,d_n\}$ is then fed into a classifier for the final prediction.\\

\noindent
\paragraph{Trade-off of ST-based approaches.}
While the shapelet transform (ST) approach provides transparency by revealing how individual shapelets influence predictions, we identify a fundamental limitation that prevents us from adopting this approach. The shapelet transform faces an inherent \textbf{trade-off} in shapelet length selection: short shapelets fail to capture temporal dependencies adequately, while long shapelets, though capable of encoding richer temporal patterns, sacrifice local interpretability by making it difficult to identify which specific shapes serve as the key patterns for instance-level predictions.
\begin{figure}[H]
\centering
\includegraphics[width=1.0\linewidth]{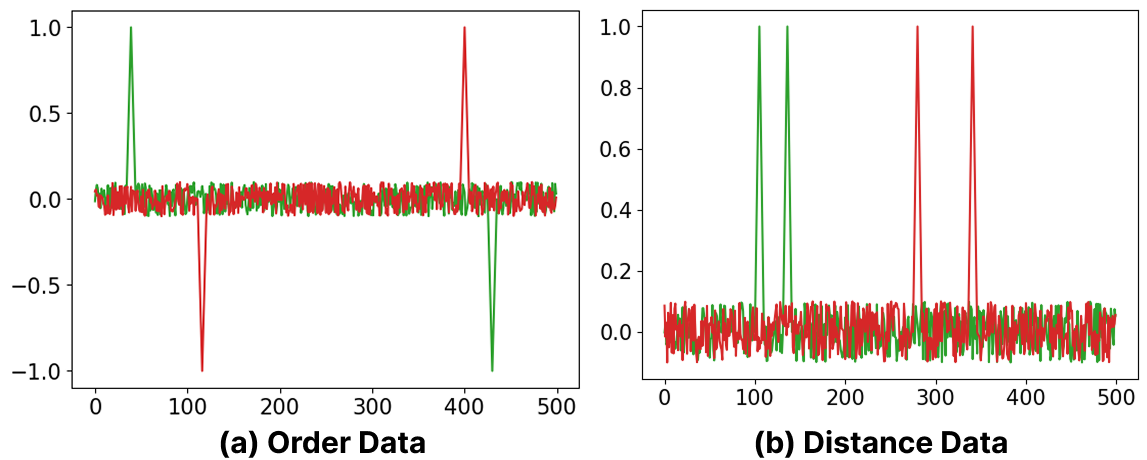}
\caption{\textbf{Synthetic dataset examples.} (Left) Order dataset. (Right) Distance dataset.}
\label{fig:placeholder}
\end{figure}
\noindent
\paragraph{Synthetic dataset for empirical test.} To empirically validate this argument, we construct two synthetic peak datasets designed to isolate different temporal decision factors: ordering and distance between peaks.
Each time series contains two peaks of fixed length (11 time steps) embedded in background noise, where peak positions are randomly sampled subject to specific constraints that test either temporal ordering or inter-peak distance as the classification criterion.
\begin{itemize}[leftmargin=*]
    \item \emph{Order} dataset, class labels are determined by the relative ordering of the two peaks.
    Specifically, one class corresponds to an up-peak followed by a down-peak (up$\rightarrow$down), while the other corresponds to a down-peak followed by an up-peak (down$\rightarrow$up).
    \item \emph{Distance} dataset, class labels are determined by the temporal distance between the two peaks.
    One class contains peaks separated by a short distance (20 time steps), while the other contains peaks separated by a long distance (50 time steps).
\end{itemize}

\begin{table}[ht]
\centering
\begin{adjustbox}{width=\linewidth}
\begin{tabular}{lcccc|c}
\toprule
\textbf{Dataset} & \textbf{LTS} & \textbf{ShapeNet}  &\textbf{ShapeConv}& \textbf{SBM} & \textbf{Ours} \\
\midrule
\textit{Distance Dataset} & 0.531& 0.507&0.4980& 0.552& \textbf{1.000}\\
\textit{Order Dataset} & 0.522& 0.525&0.534& 0.528& \textbf{1.000}\\
\bottomrule
\end{tabular}
\end{adjustbox}
\caption{\textbf{Quantitative analysis.} Classification accuracy on synthetic datasets.}
\label{tab:synthetic_results}
\end{table}

\begin{figure}[ht]
    \centering
    \includegraphics[width=1.0\linewidth]{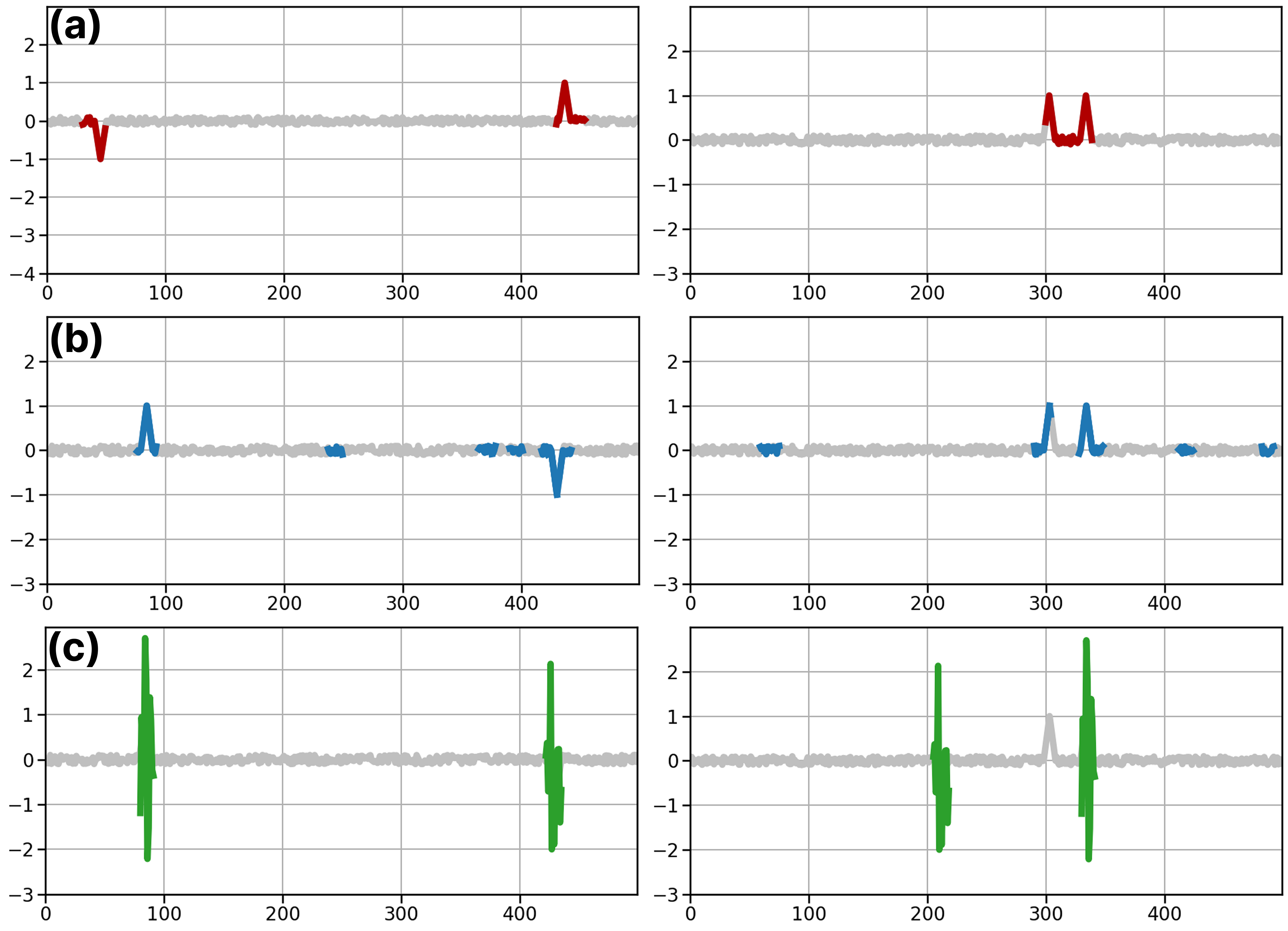}
    \caption{\textbf{Qualitative analysis.} Visualization of discovered shapelets for (a) Ours, (b) clustering-based ShapeNet, and (c) learning-based SBM.}
    \label{fig:synthetic_qualitative}
\end{figure}

\noindent
In this experiment, we evaluate whether shapelet transform-based methods can capture temporal dependencies between multiple shapelets \textbf{while preserving interpretability}.
To this end, we consider representative shapelet-aware models that rely on the shapelet transform~\cite{grabocka_LTS_KDD2014,li_shapenet_AAAI2021,qu_ShapeConv_ICLR2024,wen_InterpGN_ICLR2025}.
To explicitly expose their limitations, we fix the shapelet length of each model to the true peak length (11 time steps) and evaluate classification accuracy on the two synthetic datasets, as reported in Table~\ref{tab:synthetic_results}, together with qualitative visualizations of the discovered shapelets in Figure~\ref{fig:synthetic_qualitative}.
Despite the simplicity of the datasets, shapelet transform-based methods exhibit near-random performance, indicating their inability to capture the temporal dependencies required for correct classification.

Although these methods can attain perfect accuracy by using longer shapelets that span multiple peaks, this degrades interpretability, as they fail to identify which regions are truly important for prediction.
Notably, temporal dependencies among multiple discriminative patterns are common in real-world time-series data, amplifying the impact of this limitation in practice.
In contrast, as shown in Figure~\ref{fig:synthetic_qualitative}, our model precisely selects only the discriminative regions and achieves superior predictive performance by explicitly modeling temporal dependencies between instance-level shapelets.

\section{Experimental Details}

We reproduce all baseline methods following their official codebases and provided hyperparameter settings.\footnote{\url{https://openreview.net/forum?id=O8ouVV8PjF}}\footnote{\url{https://www.comp.hkbu.edu.hk/~csgzli/mtsc/}}\footnote{\url{https://github.com/YunshiWen/InterpretGatedNetwork}}\footnote{\url{https://github.com/ZLiu21/SoftShape}}\footnote{\url{https://github.com/rdzuo/svp-transformer}} The key characteristics of each shapelet-based method and modifications for fair comparison are summarized as follows:
\begin{itemize}[leftmargin=*]
    \item \textbf{LTS~\cite{grabocka_LTS_KDD2014}:} This method formulates shapelet discovery as a learning problem rather than an exhaustive search. It jointly optimizes shapelet patterns and a linear classifier using stochastic gradient descent, with a soft-minimum function to compute differentiable distances between time series and shapelets.
    
    \item \textbf{ShapeNet~\cite{li_shapenet_AAAI2021}:} This model addresses multivariate time series classification by embedding shapelet candidates of varying lengths into a unified representation space using dilated causal convolutions. It employs a cluster-wise triplet loss for training and selects representative shapelets through clustering, followed by an SVM classifier.
    
    \item \textbf{SVP-T~\cite{zuo2_SVPT_AAAI2023}:} This transformer-based method operates at the shape level rather than the timestamp level. It uses clustering to identify discriminative subsequences from different variables and positions, encoding them with a novel positional embedding scheme that captures variable ID, start timestamp, and end timestamp.
    
    \item \textbf{ShapeConv~\cite{qu_ShapeConv_ICLR2024}:} This method establishes a theoretical equivalence between convolution operations and shapelet transforms, treating convolutional kernels as shapelets. It introduces shaping regularization to ensure learned kernels remain interpretable as meaningful subsequence patterns.
    
    \item \textbf{SBM~\cite{wen_InterpGN_ICLR2025}:} The Shapelet Bottleneck Model constructs logical predicates using a variant of shapelet transforms to provide interpretable classification decisions based on shapelet distances.
    
    \item \textbf{InterpGN~\cite{wen_InterpGN_ICLR2025}:} This model builds upon SBM by integrating it with a deep neural network through a mixture-of-experts (MoE) gating mechanism. The gating function routes samples based on the confidence of the interpretable expert. Since the deep learning module processes the entire input time series, it cannot provide interpretability. We therefore compare only its SBM component in our experiments.

    \item \textbf{SoftShape~\cite{liu_SoftShape_ICML2025}:} This method introduces soft shape sparsification, which assigns attention-based weights to candidate subsequences based on their classification contribution scores rather than discarding non-selected shapes. It employs an MoE architecture for intra-shape pattern learning and a shared expert for inter-shape dependencies. The original architecture applies top-$k$ thresholding to select informative segments but then processes both selected and non-selected segments through shared deep networks, causing information mixing that obscures which segments truly drive classification. We use a modified version that processes only the top-$k$ segments, eliminating this mixing. The top-$k$ value is set to match the coverage ratio of INSHAPE for fair comparison.

\end{itemize}
To comprehensively evaluate INSHAPE's performance, we also compare against the full versions of InterpGN and SoftShape that leverage deep learning modules on the entire time series, as shown in Table~\ref{tab:ucr128_comparison_interpgn_softshape}.
\begin{table}[H]
\centering
\small  
\begin{tabular}{lccc}
\toprule
\multirow{2}{*}{Metric} & \multicolumn{3}{c}{UCR 128} \\
\cmidrule(lr){2-4}
 & InterpGN & SoftShape & Ours \\
\midrule
Avg. Acc & 0.7988 & 0.8400 & \textbf{0.8405} \\
\bottomrule
\end{tabular}
\caption{Comparison with full versions of InterpGN and SoftShape on UCR 128 datasets.}
\label{tab:ucr128_comparison_interpgn_softshape}
\end{table}
\noindent
Despite using deep learning modules that process the entire time series, INSHAPE achieves comparable or superior performance. This demonstrates the effectiveness of identifying statistically coherent regions and modeling their interactions, rather than relying on global time series information.


\subsection{Implementation Details}
To instantiate the selector, we adopt a Transformer encoder coupled with a lightweight MLP head, both of which are shared across all segments. This design allows the model to efficiently process variable-length segments within a single forward pass. In addition, parameter sharing significantly improves memory efficiency and acts as a regularizer, reducing the risk of overfitting and mitigating noise sensitivity—issues commonly reported in Transformer-based architectures~\cite{eldele_tslanet_ICML2024}. Furthermore, the selector is permutation-invariant by design, ensuring a consistent and reliable selection of the most informative temporal patterns at the segment level for the downstream predictor. For the predictor, we utilize an InceptionTime module~\cite{ismail_inceptiontime_DMKD2020}, which has demonstrated superior performance in time series classification tasks due to its ability to capture a wide range of temporal patterns using multi-level convolutional filters~\cite{middlehurst_backoffredux_DMKD2024}. This capability in extracting diverse patterns enables the predictor to serve as a good signal for training the selector. In particular, as the selector and predictor are jointly optimized, the selector can more effectively identify and select the segments that are most informative for the predictor.

When adapting INSHAPE to multivariate time-series data, we employ a shared selector across channels while adding a learnable channel position embedding to the input, enabling channel-specific selection strategies. The InceptionTime~\cite{ismail_inceptiontime_DMKD2020} predictor processes the full multi-channel input, with dimensionality reduced via bottleneck layers. 

All experiments are conducted on a single GPU machine\footnote{CPU: Intel Xeon Gold 6526Y (16 cores, 32 threads); GPU: NVIDIA A6000 (48GB VRAM).}.

\subsection{Model and Dataset Configurations}

\paragraph{Model Hyperparameters.}
Table~\ref{tab:hyperparams} summarizes the hyperparameter settings for INSHAPE. We use Adam optimizer~\cite{kigma_adam_ICLR2014} with He initialization~\cite{he_Heinitial_IEEE2015}. The selector and predictor are trained with different learning rates (0.0005 and 0.001, respectively) to balance their joint optimization. The coefficient $\lambda$ and PELT hyperparameter $\beta$ are tuned via grid search from \{0.1, 0.2, 0.5\} and \{0.1, 0.5, 1\}, respectively.

\begin{table}[t]
\centering
\begin{adjustbox}{width=\linewidth}
\begin{tabular}{@{}ll@{}}
\toprule
\textbf{Component} & \textbf{Hyperparameter Setting} \\
\midrule
\multicolumn{2}{@{}l}{\textbf{General Setup}} \\
\quad Initialization & He~\cite{he_Heinitial_IEEE2015} \\
\quad Optimization & Adam~\cite{kigma_adam_ICLR2014} \\
\quad Mini-batch size & \{16, 64, 512\} (UEA), \{64, 512\} (UCR) \\
\addlinespace[0.5ex]
\multicolumn{2}{@{}l}{\textbf{Training Configuration}} \\
\quad Predictor learning rate & 0.001 \\
\quad Selector learning rate & 0.0005 \\
\quad Training epochs & 500 \\
\quad Coefficient $\lambda$ & \{0.1, 0.2, 0.5\} \\
\quad PELT hyperparameter $\beta$ & \{0.1, 0.5, 1\} \\
\addlinespace[0.5ex]
\multicolumn{2}{@{}l}{\textbf{Model Architecture}} \\
\quad No. of transformer layers & 1 \\
\quad No. of inception layers & 6 \\
\quad No. of InceptionTime kernels & 3 \\
\quad Kernel size & 41 \\
\quad Bottleneck size & 32 \\
\quad Hidden dimension & 32 \\
\bottomrule
\end{tabular}
\end{adjustbox}
\caption{Hyperparameters of \textbf{INSHAPE}}
\label{tab:hyperparams}
\end{table}

\paragraph{Dataset Splits and Preprocessing.}
For the UCR datasets, we follow the train-validation-test split of 60\%-20\%-20\% as in~\cite{liu_SoftShape_ICML2025}. For the UEA datasets, we use the predefined train-test splits, following the work~\cite{wen_InterpGN_ICLR2025}. Detailed statistics of the UCR~\cite{dau_ucr_JAS2019} and UEA~\cite{bagnall_UEA_arxiv} datasets are available at the official archive\footnote{https://www.timeseriesclassification.com/dataset.php}.

\subsection{Detailed Metric Definitions}
We introduce coverage and overlap metrics to quantify the quality of local interpretations. These metrics evaluate how compactly and distinctly shapelets explain the time series. 
Let $\xv\in\R^T$ denote a time series instance and $\sv^* = \{s_1^*,\ldots,s_{M^*}^*\}$ denote the set of selected shapelets, where each shapelet $s_m^*=\{\xv_{\tau_m^s},\ldots,\xv_{\tau_m^e}\}$ corresponds to a contiguous subsequence of $\xv$. 
We define coverage and overlap as follows:
\begin{equation}
    \text{Coverage} = \frac{|\bigcup_{i=1}^{M^*}s^*_i|}{|\xv|},\quad
    \text{Overlap} = \frac{\sum_{i\neq j}|s_i\cap s_j|}{|\xv|}
\end{equation}
where $|\cdot|$ denotes the number of time points (i.e., $|\xv|=T$).
Coverage measures the proportion of the time series covered by the selected shapelets. Lower coverage indicates that the model focuses on fewer, more critical regions, providing more compact explanations.
Overlap measures the proportion of temporal positions that are covered by multiple shapelets. Lower overlap indicates that shapelets highlight distinct, non-redundant regions, making it easier to identify which specific parts contribute to the prediction.

\subsection{Algorithm Details}~\label{sup:Alorithm_Details}
The resulting population-level shapelets can be utilized for both local and global interpretations with the trained selector. For local interpretations, aligning the population-level shapelets with the regions identified by the trained selector reveals how globally important patterns manifest within individual instances. For global interpretations, we identify which population-level shapelets are close to the selected regions for each instance, and then statistically analyze these correspondences to determine which shapelets are essential for characterizing each class.
\begin{itemize}[leftmargin=*]
\item \textbf{Algorithm~\ref{alg:dtw_clustering}} details the aggregation of variable-length instance-level shapelets into population-level shapelets via DTW-based clustering. We determine the optimal number of clusters $k$ per class by maximizing silhouette scores, evaluating $k$ in the range $[5, 15]$. Additionally we define a normalized DTW metric by dividing the distance by the maximum length of the two compared shapelets. We remove a shapelet if its normalized distance to any existing one falls below the threshold $\epsilon = 0.8$.
    \item \textbf{Algorithm~\ref{alg:local_interpretation}} provides local interpretations by aligning population-level shapelets with instance-specific discriminative regions, revealing how global patterns manifest in individual instances. To preserve the characteristic shape of population-level shapelets during the interpolation and pooling process, we use DTW warping path projection. This approach aligns shapelets by following the DTW path: averaging values for compression and repeating values for stretching, thereby maintaining the original pattern morphology.
    \item \textbf{Algorithm~\ref{alg:global_interpretation}} provides global interpretations by computing usage frequency statistics of population-level shapelets across classes, identifying which shapelets are essential for characterizing each class.
\end{itemize}

\renewcommand{\algorithmicrequire}{\textbf{Input:}}
\renewcommand{\algorithmicensure}{\textbf{Output:}}

\begin{algorithm}[H]
\footnotesize
\caption{DTW-based Clustering for Population-level Shapelets}
\label{alg:dtw_clustering}
\begin{algorithmic}[1]
\Require Trained selector $\pi_\phi$, training set $\mathcal{D} = \{(\mathbf{x}^{(i)}, y^{(i)})\}_{i=1}^{N}$, number of clusters per class $K = \{k_1, \ldots, k_C\}$, similarity threshold $\epsilon$
\Ensure Population-level shapelets $\mathcal{P} = \{p_1, \ldots, p_{n_{\mathcal{P}}}\}$

\State Extract instance-level shapelets from all training instances using $\pi_\phi$:
\Statex \hspace{\algorithmicindent} $\mathcal{S}^{*} \gets \{(\sv^{*}_{i}, y^{(i)})\}_{i=1}^{N}$

\State Partition $\mathcal{S}^*$ by class labels:
\Statex \hspace{\algorithmicindent} $\mathcal{S}_c \gets \{\sv^*_i \mid (\sv^*_i, y^{(i)}) \in \mathcal{S}^*, y^{(i)} = c\}$ for $c \in \{1, \ldots, C\}$

\State $\mathcal{P} \gets \emptyset$
\For{$c = 1$ to $C$}
    \State $\{p_1, \ldots, p_{k_c}\} \gets \textsc{DTW-Kmeans}(\mathcal{S}_c, k_c)$ \Comment{Cluster centroids}
    \State $\mathcal{P} \gets \mathcal{P} \cup \{p_1, \ldots, p_{k_c}\}$
\EndFor

\State Remove redundant patterns from $\mathcal{P}$ with normalized DTW-distance $< \epsilon$
\State \Return $\mathcal{P}$ 
\end{algorithmic}
\end{algorithm}

\begin{algorithm}[t]
\footnotesize
\caption{Local Interpretation via Population-level Shapelets}
\label{alg:local_interpretation}
\begin{algorithmic}[1]
\Require Trained selector $\pi_\phi$, input instance $\mathbf{x}_{1:T}$, population-level shapelets $\mathcal{P} = \{p_1, \ldots, p_{n_\mathcal{P}}\}$.
\Ensure Interpreted instance $\tilde{\mathbf{x}}$ with discriminative regions replaced by aligned population-level shapelets

\State Extract instance-level shapelets $\sv^* = \{s_1^*, \ldots, s_{M^*}^*\}$ from $\mathbf{x}$ using $\pi_\phi$
\State Initialize aligned shapelets $\tilde{\mathcal{P}} \gets \emptyset$

\For{each $s_j^* \in \sv^*$}
    \State $p^* \gets \arg\min_{p \in \mathcal{P}} \textsc{DTW}(p, s_j^*)$ \Comment{Find closest shapelet}
    \If{$|p^*| < |s_j^*|$}
        \State $\tilde{p}_j \gets \textsc{Interpolate}(p^*, s_j^*)$ \Comment{Upsample}
    \ElsIf{$|p^*| > |s_j^*|$}
        \State $\tilde{p}_j \gets \textsc{Pool}(p^*, s_j^*)$ \Comment{Downsample}
    \Else
        \State $\tilde{p}_j \gets p^*$
    \EndIf
    \State $\tilde{\mathcal{P}} \gets \tilde{\mathcal{P}} \cup \{\tilde{p}_j\}$
\EndFor

\State $\tilde{\mathbf{x}} \gets \textsc{Align}(\mathbf{x}, \sv^*, \tilde{\mathcal{P}})$ \Comment{Replace $\sv^*$ with $\tilde{\mathcal{P}}$}
\State \Return $\tilde{\mathbf{x}}$
\end{algorithmic}
\end{algorithm}

\begin{algorithm}[t]
\caption{Global Interpretation via Population-level Shapelets}
\label{alg:global_interpretation}
\begin{algorithmic}[1]
\Require Trained selector $\pi_\phi$, partitioned instance-level shapelets $\mathcal{S}^* = \{\mathcal{S}_1, \ldots, \mathcal{S}_C\}$, population-level shapelets $\mathcal{P} = \{p_1, \ldots, p_{n_{\mathcal{P}}}\}$
\Ensure Usage frequency matrix $\mathbf{U} \in \mathbb{R}^{C \times n_{\mathcal{P}}}$ where $u_{c,k}$ denotes the frequency of $p_k$ in class $c$

\State Initialize $\mathbf{U} \gets \mathbf{0}^{C \times n_{\mathcal{P}}}$

\For{$c = 1$ to $C$}
    \For{each $\sv^* \in \mathcal{S}_c$}
        \For{each $s_j \in \sv^*$}
            \State $k \gets \arg\min_{k \in \{1, \dots, n_\mathcal{P}\}} \textsc{DTW}(p_k, s_j)$ \Comment{Find closest shapelet}
            \State $u_{c, k} \gets u_{c, k} + 1$
        \EndFor
    \EndFor
\EndFor

\State \Return $\mathbf{U}$
\end{algorithmic}
\end{algorithm}

\begin{figure}[t]
    \centering
    \includegraphics[width=1.0\linewidth]{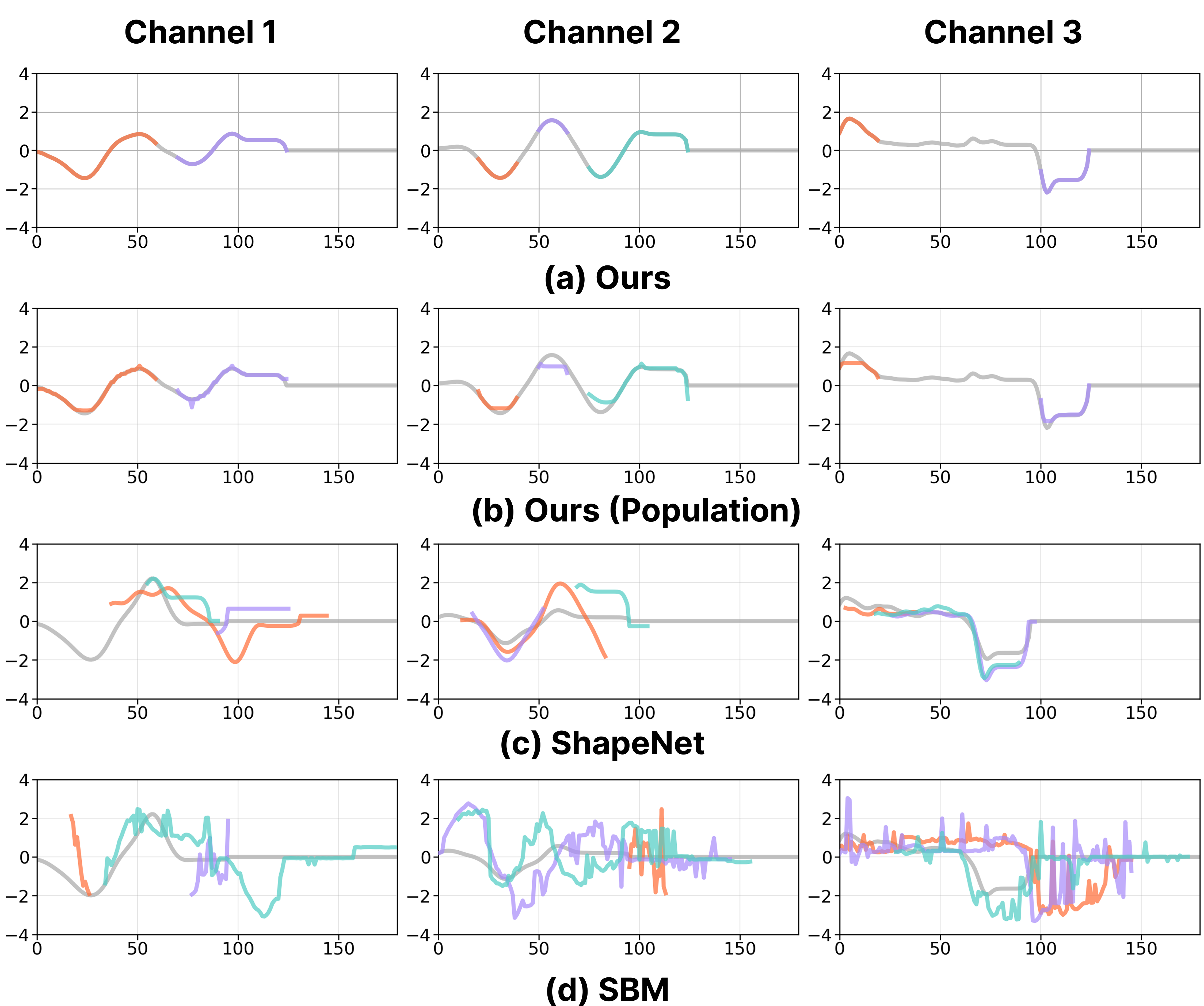}
    \caption{\textbf{Qualitative comparison of local interpretations on a multivariate time series from CharacterTrajectories (UEA).} 
    Each column represents a channel (pen trajectory dimension), and colored segments indicate discovered shapelets. Gray regions denote non-selected parts of the time series.}
    \label{fig:multivariate_qualitative}
\end{figure}

\section{Additional Results}
INSHAPE is an interpretable-by-design model that provides both local and global explanations. 
In this section, we present additional qualitative results on multivariate time series, computational complexity analysis, and full results corresponding to Table~2 in the main paper.

\subsection{Local Interpretation on Multivariate Time Series}
Figure~\ref{fig:multivariate_qualitative} compares local interpretations on a CharacterTrajectories sample from the UEA archive, which consists of three channels representing pen movement trajectories.
In (a), INSHAPE identifies instance-level shapelets that capture distinct, non-overlapping discriminative segments across channels.
In (b), we show local interpretation via population-level shapelets, where population-level shapelets are matched to their corresponding instance-level shapelets based on FastDTW distance.

In contrast, the baseline methods exhibit notable limitations. ShapeNet (c) produces shapelets with substantial overlap across different patterns, making it difficult to distinguish which temporal regions are truly discriminative. SBM (d) generates fragmented and noisy shapelets that fail to capture coherent temporal structures, resulting in interpretations that are difficult to understand. These comparisons highlight that INSHAPE provides clearer, more coherent local explanations suitable for multivariate time-series analysis.

\subsection{Full Results of Global Interpretation}
In the main paper, we presented quantitative analysis using the usage frequency matrix $\mathbf{U}$ for the ECG5000 dataset, with detailed interpretation for classes 0 and 4. Here, we provide the full results across all five classes.
\begin{figure*}[!t]
    \centering
    \includegraphics[width=1.0\linewidth]{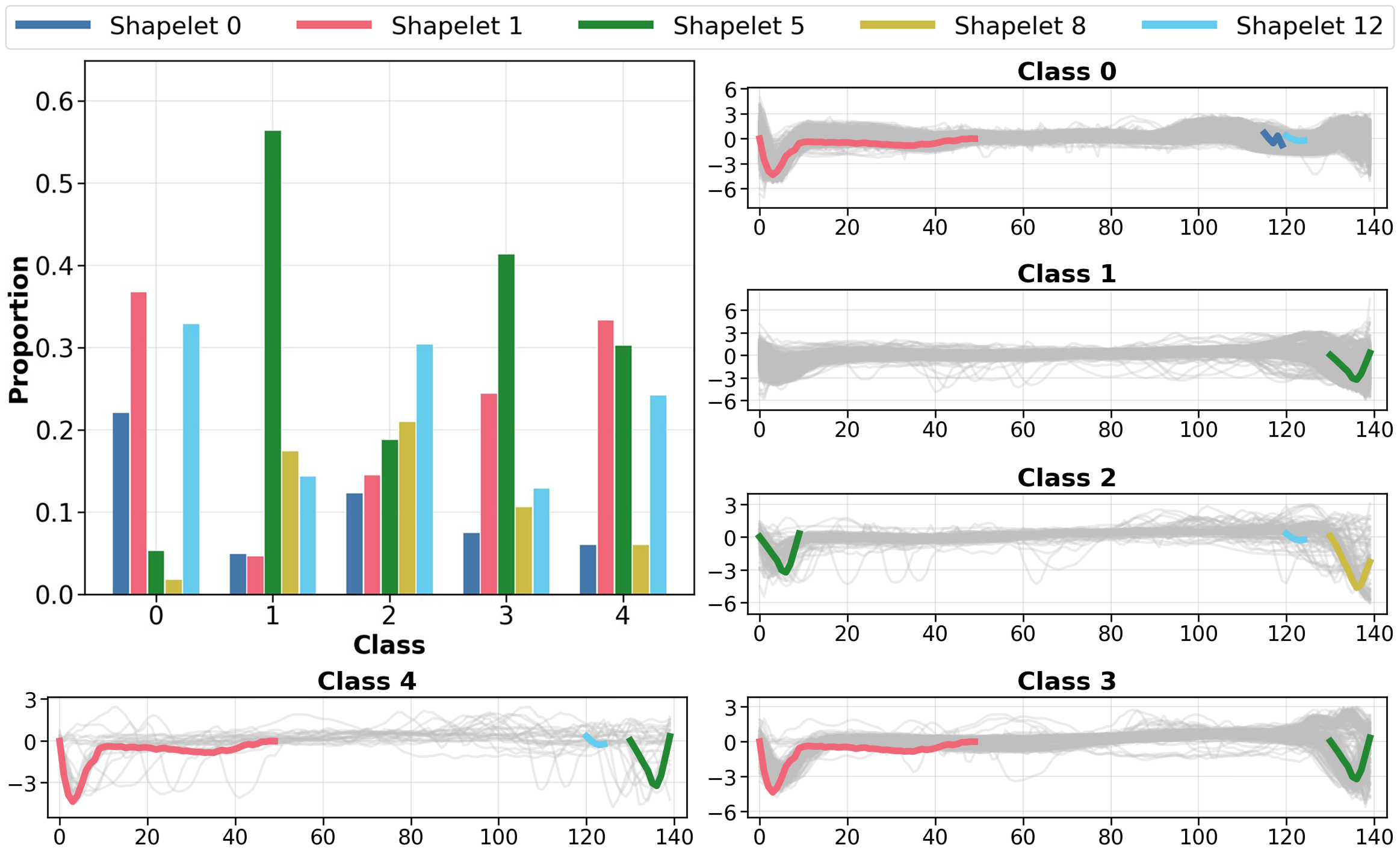}
    \caption{\textbf{Full global interpretation results on ECG5000.} 
    (Left) Class-wise proportion of the top-5 population-level shapelets. 
    (Right) Overlay of the corresponding shapelets on time-series instances for each class, with gray lines representing individual samples.}
    \label{fig:global_interpretation_full_ECG5000}
\end{figure*}
Figure~\ref{fig:global_interpretation_full_ECG5000} shows the class-wise proportion of each population-level shapelet (Left) and their overlay on representative instances per class (Right).
Class 1 is predominantly characterized by Shapelet 5, which consistently appears in the tail region across instances.
In contrast, Class 2 exhibits a more balanced distribution across multiple shapelets ($p_5$, $p_8$, $p_{12}$), requiring consideration of temporal dependency for discrimination with other classes.
Classes 3 and 4 share similar shapelet distributions, with Shapelets 1 and 5 being dominant in both. This behavior parallels tree-based models~\cite{koegh_Shapelet_KDD2009,mueen_logicalshapelet_KDD2011}, where features are selected to maximize information gain—suggesting that INSHAPE implicitly learns to identify shapelets that are most informative for distinguishing between classes.
\begin{figure*}[t]
    \centering
    \includegraphics[width=1.0\linewidth]{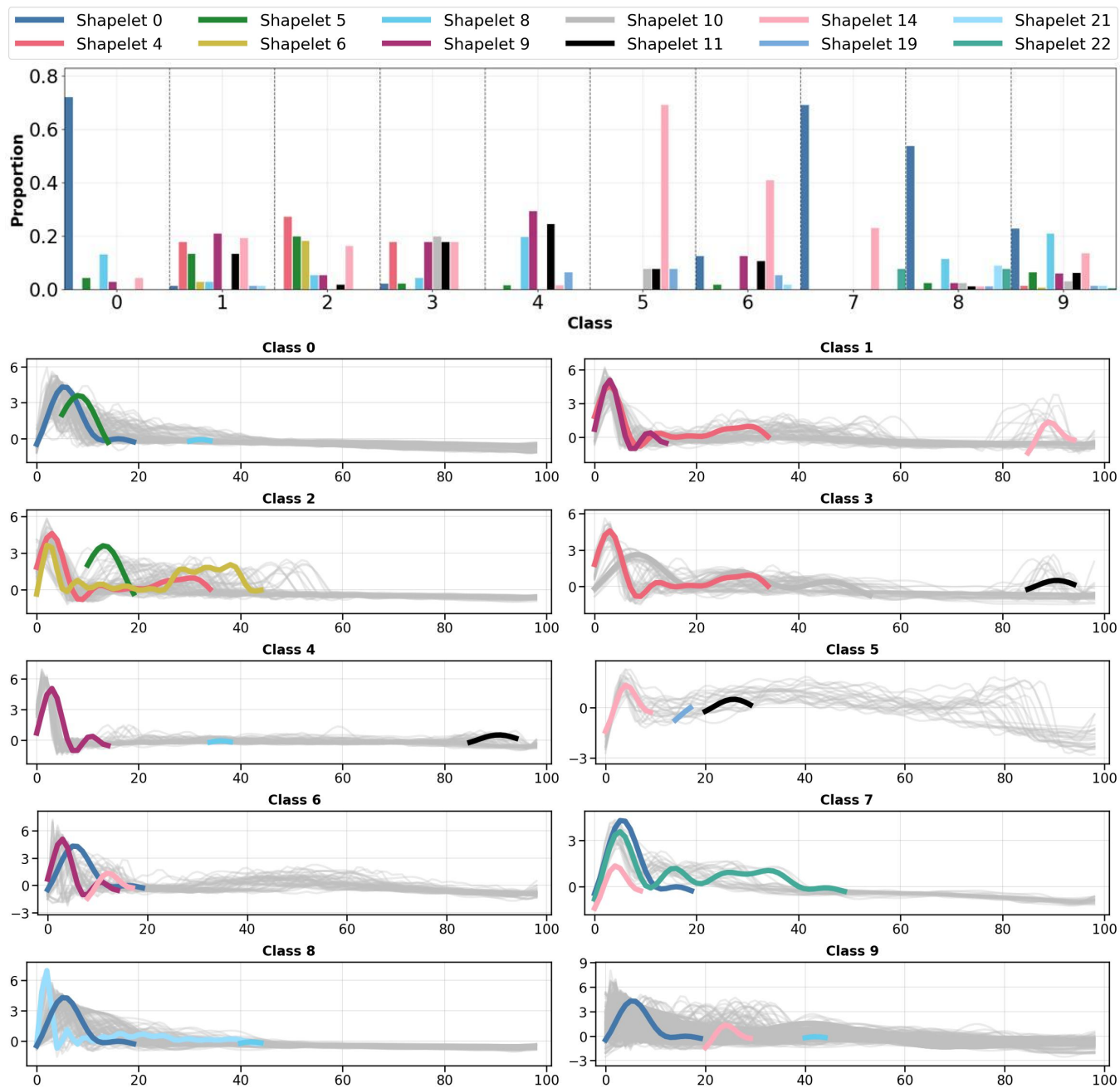}
    \caption{\textbf{Global interpretation results on MedicalImages.} (Top) Class-wise proportion of the top-3 population-level shapelets extracted from the frequency matrix $\mathbf{U}$. (Bottom) Overlay of corresponding shapelets on time-series instances for each class. Gray lines represent individual samples.}
    \label{fig:global_interpretation_full_MedicalImages}
\end{figure*}
Figure~\ref{fig:global_interpretation_full_MedicalImages} shows the top-3 frequent shapelets for each class in the MedicalImages dataset. Similar to the ECG5000 analysis, the proportion bar plot derived from the frequency matrix $\mathbf{U}$ enables intuitive class-level interpretation. Class 2 exhibits a more balanced distribution across Shapelets $p_4$, $p_5$, and $p_6$, suggesting the temporal dependency for discriminating class.

\subsection{Computational Cost Analysis}
We investigate the computational cost of INSHAPE by measuring the training time on the ECG5000 dataset.
Table~\ref{tab:computational_cost} reports the average training time per epoch.
The results show that INSHAPE maintains practical training efficiency compared with existing shapelet-based baselines.

\begin{table}[!t]
    \centering
    \small
    \begin{tabular}{lcccc}
        \toprule
        & LTS & ShapeConv & SBM & Ours \\
        \midrule
        Train (s) & 0.74 & 1.45 & 0.67 & 1.17 \\
        \bottomrule
    \end{tabular}
    \caption{Computational cost comparison in seconds. Training time is measured per epoch on ECG5000.}
    \label{tab:computational_cost}
\end{table}

\paragraph{Inference Cost in a Real-World Scenario.}
We further evaluate the inference efficiency of INSHAPE in an EEG-based sleep-stage classification.
Following standard sleep-stage prediction protocols, each input sample corresponds to a 30-second EEG segment sampled at 128 Hz~\cite{perslev_Usleep_2021}, resulting in 3,840 time points per inference.
Under this setting, INSHAPE requires an average inference time of 0.4913 seconds per sample and a peak GPU memory usage of 19.85 MB, as reported in Table~\ref{tab:inference_cost}.
These results suggest that INSHAPE is computationally feasible for practical deployment scenarios, including frequent dynamic prediction and resource-constrained environments.

\begin{table}[!t]
    \centering
    \small
    \begin{tabular}{lc}
        \toprule
        Metric & Value \\
        \midrule
        Average inference time (s) & 0.4913 \\
        Peak GPU memory (MB) & 19.85 \\
        \bottomrule
    \end{tabular}
    \caption{Average per-sample inference time and peak GPU memory usage of INSHAPE on the EEG dataset, where each sample consists of a 30-second segment sampled at 128 Hz.}
    \label{tab:inference_cost}
\end{table}

\subsection{Domain-Grounded Analysis on ECG5000}
\begin{figure*}[!t]
    \centering
    \includegraphics[width=0.9\textwidth]{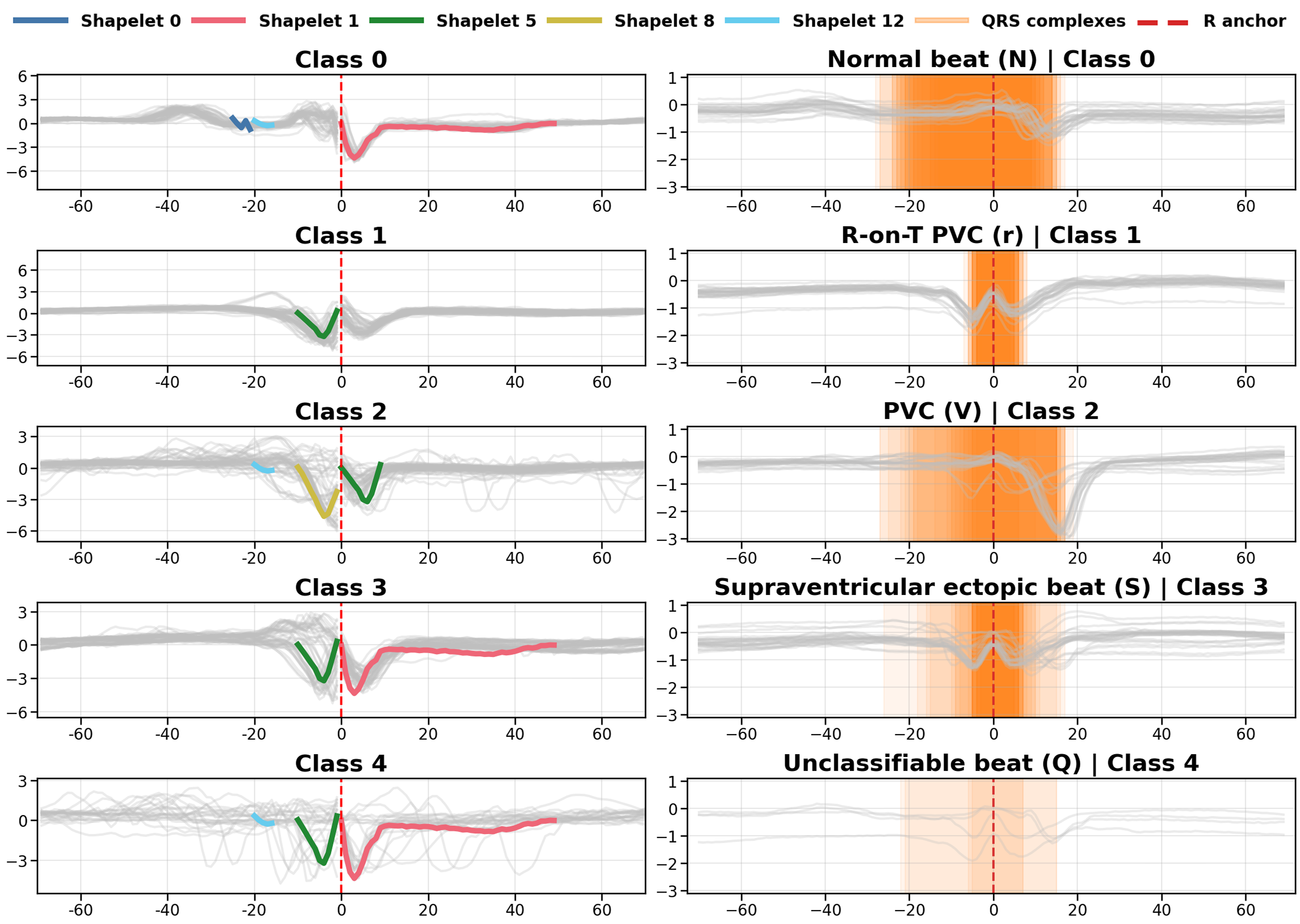}
    \caption{\textbf{Domain-grounded analysis on ECG5000.}
    Population-level shapelets discovered by INSHAPE are overlaid on ECG5000 heartbeat instances, together with the corresponding clinical reference regions based on the QRS complex.
    The alignment between the discovered shapelets and the QRS regions indicates that INSHAPE captures discriminative temporal patterns consistent with ECG-domain knowledge.}
    \label{fig:ecg_clinical_reference}
\end{figure*}
To further examine whether the shapelets discovered by INSHAPE are consistent with domain knowledge, we conduct an additional analysis on ECG5000.
ECG5000 is constructed from the BIDMC Congestive Heart Failure Database by extracting and normalizing fixed-length heartbeat segments~\cite{dau_ucr_JAS2019}.
The dataset categorizes heartbeats into five classes: normal beat, R-on-T premature ventricular contraction, premature ventricular contraction, supraventricular premature or ectopic beat, and unclassifiable beat.

In electrocardiogram analysis, the QRS complex, which corresponds to ventricular depolarization, is one of the most important morphological components for distinguishing normal beats from ectopic or abnormal beats~\cite{diffdiag_abedin_JA2021}.
Motivated by this clinical observation, we construct a domain-grounded reference region by aligning class-annotated beats from the original database with respect to the R-peak and identifying the corresponding Q-to-S interval.
This interval provides a clinically meaningful reference for assessing whether the discovered shapelets are localized around diagnostically relevant waveform regions.

Figure~\ref{fig:ecg_clinical_reference} shows the population-level shapelets discovered by INSHAPE and the corresponding QRS reference regions.
The selected shapelets are consistently localized within or near the QRS complex, indicating that INSHAPE captures temporal patterns that are well aligned with established ECG morphology.
This result provides additional evidence that the proposed instance-level shapelet discovery process can identify discriminative subsequences that are not only predictive but also meaningful from a domain perspective.

It is important to note that ectopic beats often exhibit substantial heterogeneity across patients, including variations in morphology, timing, and cardiac-cycle position.
Therefore, a fixed population-level pattern may not fully capture the patient-specific characteristics required for reliable interpretation.
By selecting shapelets at the instance level, INSHAPE can adaptively identify discriminative regions tailored to each individual heartbeat while still supporting population-level analysis through the aggregation procedure described in Section~\ref{sup:Alorithm_Details}.
This property is particularly useful in clinical time-series analysis, where patient-specific temporal patterns may carry important diagnostic information.


\subsection{Interpretability of Contiguous Instance-Level Shapelets}
INSHAPE discovers instance-level shapelets as contiguous and non-overlapping temporal segments.
This design choice improves interpretability because each selected region can be understood as a coherent temporal unit rather than as a fragmented collection of isolated timestamps.
From a perceptual perspective, this is consistent with the Gestalt principle of connectedness, where connected visual elements are naturally perceived as belonging to the same group~\cite{rock_gestalt_SA1990}.
In time-series interpretation, contiguous segments therefore provide a more intuitive explanation of which temporal regions contribute to the prediction.

This property also distinguishes INSHAPE from attribution-based explanations that may highlight scattered timestamps across the input sequence.
Although such fragmented explanations can indicate point-wise importance, they are often difficult to interpret as meaningful temporal patterns.
In contrast, the contiguous shapelets discovered by INSHAPE correspond to local waveform structures, making them easier to inspect, compare, and aggregate across instances.
Together with the non-overlapping selection constraint, this leads to compact explanations that identify distinct discriminative regions without redundant temporal coverage.
\newpage
\clearpage

\onecolumn
\footnotesize
\setlength\LTleft{\fill}
\setlength\LTright{\fill}
\setlength{\tabcolsep}{3pt}
\renewcommand{\arraystretch}{1.05}

\begin{longtable}{l|ccccc|cccccc|c}
\toprule
dataset & \#train & \#test & \#total & length & \#class & LTS & ShapeNet & SVP-T & ShapeConv & SBM & SoftShape & Ours \\
\midrule
\endfirsthead

\toprule
dataset & \#train & \#test & \#total & length & \#class & LTS & ShapeNet & SVP-T & ShapeConv & SBM & SoftShape & Ours \\
\midrule
\endhead

\midrule
\multicolumn{13}{r}{\small\itshape Continued on next page}\\
\endfoot

\bottomrule
\endlastfoot

ACSF1 & 100 & 100 & 200 & 1460 & 10 & 0.5000 & 0.6450 & 0.5500 & 0.7000 & 0.5750 & \textbf{0.8050} & 0.7450 \\
Adiac & 390 & 391 & 781 & 176 & 37 & 0.1410 & 0.4443 & 0.1274 & 0.7070 & 0.2551 & \textbf{0.7682} & 0.7183 \\
AllGestureWiimoteX & 300 & 700 & 1000 & Vary & 10 & 0.1340 & 0.4670 & 0.6100 & 0.5700 & 0.3880 & 0.2160 & \textbf{0.8120} \\
AllGestureWiimoteY & 300 & 700 & 1000 & Vary & 10 & 0.1140 & 0.4670 & 0.7150 & 0.6350 & 0.5050 & 0.4950 & \textbf{0.8400} \\
AllGestureWiimoteZ & 300 & 700 & 1000 & Vary & 10 & 0.1370 & 0.4810 & 0.4750 & 0.5700 & 0.4930 & 0.2660 & \textbf{0.7790} \\
ArrowHead & 36 & 175 & 211 & 251 & 3 & 0.5762 & \textbf{0.9529} & 0.5116 & 0.7910 & 0.7381 & 0.9051 & 0.9048 \\
Beef & 30 & 30 & 60 & 470 & 5 & 0.3167 & \textbf{0.8556} & 0.3333 & 0.4170 & 0.3167 & 0.8333 & 0.6667 \\
BeetleFly & 20 & 20 & 40 & 512 & 2 & \textbf{0.9250} & 0.5667 & 0.7500 & 0.6250 & \textbf{0.9250} & 0.8750 & 0.8250 \\
BirdChicken & 20 & 20 & 40 & 512 & 2 & 0.7500 & 0.7750 & 0.5000 & 0.8750 & 0.6750 & 0.7750 & \textbf{1.0000} \\
BME & 30 & 150 & 180 & 128 & 3 & 0.7111 & 0.7873 & 0.8611 & 0.9440 & 0.8889 & 0.9889 & \textbf{1.0000} \\
Car & 60 & 60 & 120 & 577 & 4 & 0.6333 & 0.8250 & 0.4167 & 0.7500 & 0.6917 & \textbf{0.8917} & 0.8417 \\
CBF & 30 & 900 & 930 & 128 & 3 & 0.9978 & 0.9400 & 0.9946 & \textbf{1.0000} & \textbf{1.0000} & \textbf{1.0000} & \textbf{1.0000} \\
Chinatown & 20 & 343 & 363 & 24 & 2 & 0.9288 & \textbf{0.9989} & 0.9178 & 0.9040 & 0.9315 & 0.9724 & 0.9722 \\
ChlorineConcentration & 467 & 3840 & 4307 & 166 & 3 & 0.5517 & 0.7833 & 0.6763 & 0.5600 & 0.5977 & 0.6336 & \textbf{0.9865} \\
CinCECGTorso & 40 & 1380 & 1420 & 1639 & 4 & 0.9754 & 0.9505 & 0.9401 & 0.9610 & 0.9972 & \textbf{0.9993} & 0.9894 \\
Coffee & 28 & 28 & 56 & 286 & 2 & 0.7818 & 0.5579 & 0.5833 & 0.9170 & 0.4545 & \textbf{1.0000} & \textbf{1.0000} \\
Computers & 250 & 250 & 500 & 720 & 2 & 0.6180 & \textbf{0.8901} & 0.6400 & 0.6300 & 0.6160 & 0.7120 & 0.8500 \\
CricketX & 390 & 390 & 780 & 300 & 12 & 0.4974 & \textbf{0.9478} & 0.6538 & 0.5900 & 0.6538 & 0.7718 & 0.8051 \\
CricketY & 390 & 390 & 780 & 300 & 12 & 0.4397 & \textbf{0.9667} & 0.6410 & 0.6920 & 0.6808 & 0.7346 & 0.7885 \\
CricketZ & 390 & 390 & 780 & 300 & 12 & 0.5128 & 0.6560 & 0.6603 & 0.6220 & 0.6782 & 0.7936 & \textbf{0.8526} \\
Crop & 7200 & 16800 & 24000 & 46 & 24 & 0.4718 & 0.4910 & \textbf{0.7565} & 0.6130 & 0.5037 & 0.7003 & 0.7472 \\
DiatomSizeReduction & 16 & 306 & 322 & 345 & 4 & 0.5815 & 0.4718 & 0.7846 & \textbf{1.0000} & 0.8646 & \textbf{1.0000} & 0.9844 \\
DistalPhalanxOutlineAgeGroup & 400 & 139 & 539 & 80 & 3 & 0.7370 & \textbf{0.9276} & 0.7315 & 0.7780 & 0.7944 & 0.7736 & 0.8519 \\
DistalPhalanxOutlineCorrect & 600 & 276 & 876 & 80 & 2 & 0.6354 & \textbf{0.9750} & 0.7784 & 0.7840 & 0.7771 & 0.6587 & 0.8114 \\
DistalPhalanxTW & 400 & 139 & 539 & 80 & 6 & 0.6944 & 0.4795 & 0.7315 & 0.7040 & \textbf{0.7556} & 0.7551 & 0.7236 \\
DodgerLoopDay & 78 & 80 & 158 & 288 & 7 & 0.3875 & 0.5046 & 0.1250 & 0.5310 & 0.5000 & \textbf{0.6139} & 0.5367 \\
DodgerLoopGame & 20 & 138 & 158 & 288 & 2 & 0.8000 & \textbf{0.9938} & 0.4688 & 0.9060 & 0.7938 & 0.8671 & 0.8155 \\
DodgerLoopWeekend & 20 & 138 & 158 & 288 & 2 & 0.9500 & 0.8014 & 0.7188 & 0.9690 & 0.9625 & \textbf{0.9808} & 0.9490 \\
Earthquakes & 322 & 139 & 461 & 512 & 2 & 0.8000 & 0.7216 & 0.7634 & 0.7960 & \textbf{0.8022} & 0.7983 & 0.7983 \\
ECG200 & 100 & 100 & 200 & 96 & 2 & 0.7300 & 0.7660 & 0.7500 & 0.6750 & 0.8800 & \textbf{0.9100} & 0.8200 \\
ECG5000 & 500 & 4500 & 5000 & 140 & 5 & 0.9332 & \textbf{0.9759} & 0.9590 & 0.9490 & 0.9416 & 0.9516 & 0.9600 \\
ECGFiveDays & 23 & 861 & 884 & 136 & 2 & 0.8960 & 0.7644 & \textbf{1.0000} & 0.7910 & \textbf{1.0000} & \textbf{1.0000} & \textbf{1.0000} \\
ElectricDevices & 8926 & 7711 & 16637 & 96 & 7 & 0.7363 & \textbf{0.9744} & 0.8630 & 0.8300 & 0.8001 & 0.8019 & 0.8076 \\
EOGHorizontalSignal & 362 & 362 & 724 & 1250 & 12 & 0.4221 & 0.4433 & \textbf{0.7517} & 0.6830 & 0.6510 & 0.7432 & 0.7224 \\
EOGVerticalSignal & 362 & 362 & 724 & 1250 & 12 & 0.4566 & 0.7588 & 0.6552 & 0.4550 & 0.5821 & 0.7029 & \textbf{0.7724} \\
EthanolLevel & 504 & 500 & 1004 & 1751 & 4 & 0.2537 & \textbf{0.9222} & 0.2736 & 0.7260 & 0.2557 & 0.4152 & 0.8600 \\
FaceAll & 560 & 1690 & 2250 & 131 & 14 & 0.4271 & 0.8100 & \textbf{0.9911} & 0.9330 & 0.9653 & 0.6933 & 0.9609 \\
FaceFour & 24 & 88 & 112 & 350 & 4 & 0.6435 & 0.9294 & 0.7391 & 0.8700 & 0.8696 & 0.9462 & \textbf{0.9565} \\
FacesUCR & 200 & 2050 & 2250 & 131 & 14 & 0.8978 & 0.9910 & \textbf{0.9933} & 0.8800 & 0.9800 & 0.9142 & 0.9542 \\
FiftyWords & 450 & 455 & 905 & 270 & 50 & 0.4552 & 0.5733 & 0.5359 & 0.5750 & \textbf{0.7768} & 0.5768 & 0.6928 \\
Fish & 175 & 175 & 350 & 463 & 7 & 0.5229 & 0.7079 & 0.6286 & 0.8710 & 0.7486 & \textbf{0.9171} & 0.8571 \\
FordA & 3601 & 1320 & 4921 & 500 & 2 & 0.9238 & 0.5399 & 0.9269 & 0.9010 & 0.9138 & \textbf{0.9441} & 0.8632 \\
FordB & 3636 & 810 & 4446 & 500 & 2 & 0.9082 & 0.7983 & 0.9090 & 0.9100 & 0.9190 & \textbf{0.9262} & 0.9135 \\
FreezerRegularTrain & 150 & 2850 & 3000 & 301 & 2 & 0.7650 & 0.6270 & \textbf{0.9983} & 0.5000 & 0.8473 & 0.9343 & 0.9433 \\
FreezerSmallTrain & 28 & 2850 & 2878 & 301 & 2 & 0.7642 & 0.3188 & \textbf{0.9965} & 0.5000 & 0.8660 & 0.9840 & 0.9600 \\
Fungi & 18 & 186 & 204 & 201 & 18 & 0.7122 & 0.7720 & 0.4390 & 0.9760 & 0.9220 & \textbf{1.0000} & \textbf{1.0000} \\
GestureMidAirD1 & 208 & 130 & 338 & Vary & 26 & 0.3853 & \textbf{0.9818} & 0.3971 & 0.4710 & 0.5794 & 0.6594 & 0.6212 \\
GestureMidAirD2 & 208 & 130 & 338 & Vary & 26 & 0.2824 & \textbf{0.7533} & 0.2353 & 0.1030 & 0.4118 & 0.5297 & 0.4524 \\
GestureMidAirD3 & 208 & 130 & 338 & Vary & 26 & 0.1618 & \textbf{0.5923} & 0.1471 & 0.2500 & 0.2265 & 0.3164 & 0.2250 \\
GesturePebbleZ1 & 132 & 172 & 304 & Vary & 6 & 0.7082 & 0.8057 & 0.7213 & 0.8850 & 0.9148 & \textbf{0.9769} & 0.8650 \\
GesturePebbleZ2 & 146 & 158 & 304 & Vary & 6 & 0.6951 & 0.7602 & 0.5738 & 0.9020 & 0.9148 & \textbf{0.9639} & 0.8719 \\
GunPoint & 50 & 150 & 200 & 150 & 2 & 0.7300 & 0.6273 & 0.8000 & 0.9750 & 0.9600 & 0.9850 & \textbf{1.0000} \\
GunPointAgeSpan & 135 & 316 & 451 & 150 & 2 & 0.5756 & 0.7769 & 0.9231 & \textbf{0.9780} & 0.9178 & 0.9690 & 0.9024 \\
GunPointMaleVersusFemale & 135 & 316 & 451 & 150 & 2 & 0.6133 & 0.9337 & \textbf{1.0000} & 0.9780 & 0.9867 & 0.9823 & \textbf{1.0000} \\
GunPointOldVersusYoung & 136 & 315 & 451 & 150 & 2 & 0.7244 & 0.9218 & \textbf{1.0000} & 0.9230 & 0.9133 & 0.9534 & 0.9690 \\
Ham & 109 & 105 & 214 & 431 & 2 & 0.5442 & \textbf{1.0000} & 0.5814 & 0.8370 & 0.5581 & 0.7759 & 0.8411 \\
HandOutlines & 1000 & 370 & 1370 & 2709 & 2 & 0.8591 & 0.5091 & 0.8102 & 0.8140 & 0.8942 & \textbf{0.9022} & 0.8569 \\
Haptics & 155 & 308 & 463 & 1092 & 5 & 0.3699 & 0.4173 & 0.3871 & 0.4410 & 0.4624 & 0.4860 & \textbf{0.5161} \\
Herring & 64 & 64 & 128 & 512 & 2 & 0.6154 & 0.2571 & 0.6154 & 0.6150 & \textbf{0.6231} & 0.5615 & 0.5871 \\
HouseTwenty & 40 & 119 & 159 & 2000 & 2 & 0.5250 & 0.7728 & 0.9062 & 0.8440 & 0.8500 & 0.9373 & \textbf{0.9498} \\
InlineSkate & 100 & 550 & 650 & 1882 & 7 & 0.2462 & \textbf{0.7434} & 0.3231 & 0.5620 & 0.3569 & 0.5046 & 0.6677 \\
InsectEPGRegularTrain & 62 & 249 & 311 & 601 & 3 & 0.7226 & 0.9400 & \textbf{1.0000} & 0.8730 & 0.8645 & 0.9871 & \textbf{1.0000} \\
InsectEPGSmallTrain & 17 & 249 & 266 & 601 & 3 & 0.6642 & 0.9468 & \textbf{1.0000} & 0.7410 & 0.7623 & 0.9887 & 0.9811 \\
InsectWingbeatSound & 220 & 1980 & 2200 & 256 & 11 & 0.6136 & 0.5404 & 0.5727 & 0.6520 & 0.6791 & \textbf{0.6995} & 0.6591 \\
ItalyPowerDemand & 67 & 1029 & 1096 & 24 & 2 & 0.9616 & 0.6099 & 0.9727 & 0.9640 & 0.9598 & 0.9316 & \textbf{0.9772} \\
LargeKitchenAppliances & 375 & 375 & 750 & 720 & 3 & 0.3507 & \textbf{0.9911} & 0.8400 & 0.7600 & 0.6840 & 0.9227 & 0.9600 \\
Lightning2 & 60 & 61 & 121 & 637 & 2 & 0.7167 & 0.8402 & 0.6400 & 0.8000 & 0.7500 & 0.7933 & \textbf{0.8750} \\
Lightning7 & 70 & 73 & 143 & 319 & 7 & 0.5931 & 0.6682 & 0.3448 & 0.4140 & 0.8207 & 0.6926 & \textbf{0.8621} \\
Mallat & 55 & 2345 & 2400 & 1024 & 8 & 0.9500 & 0.8620 & 0.9833 & 0.9850 & 0.9833 & 0.9796 & \textbf{0.9942} \\
Meat & 60 & 60 & 120 & 448 & 3 & 0.3333 & 0.4816 & 0.5000 & 0.7500 & 0.3333 & 0.9833 & \textbf{1.0000} \\
MedicalImages & 381 & 760 & 1141 & 99 & 10 & 0.6614 & 0.6252 & 0.7467 & 0.5150 & 0.6886 & 0.7142 & \textbf{0.7686} \\
MelbournePedestrian & 1194 & 2439 & 3633 & 24 & 10 & 0.6850 & \textbf{0.9725} & 0.8583 & 0.8460 & 0.8459 & 0.7947 & 0.8583 \\
MiddlePhalanxOutlineAgeGroup & 400 & 154 & 554 & 80 & 3 & 0.7135 & \textbf{0.8107} & 0.7207 & 0.7390 & 0.7568 & 0.7472 & 0.7255 \\
MiddlePhalanxOutlineCorrect & 600 & 291 & 891 & 80 & 2 & 0.6225 & 0.3431 & 0.6648 & 0.6540 & 0.6472 & 0.6442 & \textbf{0.7980} \\
MiddlePhalanxTW & 399 & 154 & 553 & 80 & 6 & 0.6036 & 0.5043 & 0.5946 & 0.6220 & 0.6216 & 0.4549 & \textbf{0.6667} \\
MixedShapesRegularTrain & 500 & 2425 & 2925 & 1024 & 5 & 0.8503 & 0.5337 & 0.9419 & 0.9330 & 0.9032 & 0.8335 & \textbf{0.9709} \\
MixedShapesSmallTrain & 100 & 2425 & 2525 & 1024 & 5 & 0.7711 & 0.8586 & 0.9347 & 0.8120 & 0.8962 & 0.8772 & \textbf{0.9596} \\
MoteStrain & 20 & 1252 & 1272 & 84 & 2 & 0.9388 & 0.8796 & 0.9451 & 0.9370 & \textbf{0.9498} & 0.9450 & 0.9418 \\
NonInvasiveFetalECGThorax1 & 1800 & 1965 & 3765 & 750 & 42 & 0.4005 & 0.5377 & 0.8898 & 0.8310 & 0.7424 & \textbf{0.8961} & 0.8853 \\
NonInvasiveFetalECGThorax2 & 1800 & 1965 & 3765 & 750 & 42 & 0.4518 & 0.9352 & 0.8911 & 0.8910 & 0.8181 & \textbf{0.9410} & 0.8898 \\
OliveOil & 30 & 30 & 60 & 570 & 4 & 0.4167 & 0.6467 & 0.4167 & 0.4170 & 0.4167 & 0.4167 & \textbf{0.6667} \\
OSULeaf & 200 & 242 & 442 & 427 & 6 & 0.5910 & 0.4762 & 0.5955 & 0.6740 & 0.7416 & 0.7965 & \textbf{0.8409} \\
PhalangesOutlinesCorrect & 1800 & 858 & 2658 & 80 & 2 & 0.6406 & 0.6860 & 0.6955 & \textbf{0.8050} & 0.6910 & 0.6949 & 0.7641 \\
Phoneme & 214 & 1896 & 2110 & 1024 & 39 & 0.1668 & \textbf{0.9788} & 0.2749 & 0.1680 & 0.2815 & 0.2028 & 0.3607 \\
PickupGestureWiimoteZ & 50 & 50 & 100 & Vary & 10 & 0.5600 & \textbf{0.9583} & 0.2000 & 0.5000 & 0.5600 & 0.7500 & 0.7700 \\
PigAirwayPressure & 104 & 208 & 312 & 2000 & 52 & 0.0254 & 0.6161 & 0.1270 & \textbf{0.9050} & 0.0254 & 0.0417 & 0.2581 \\
PigArtPressure & 104 & 208 & 312 & 2000 & 52 & 0.2000 & 0.7297 & 0.4762 & 0.8570 & 0.6571 & 0.6940 & \textbf{1.0000} \\
PigCVP & 104 & 208 & 312 & 2000 & 52 & 0.1175 & 0.7418 & 0.5079 & 0.6350 & 0.4317 & 0.8814 & \textbf{0.9516} \\
PLAID & 537 & 537 & 1074 & Vary & 11 & 0.1981 & \textbf{0.7353} & 0.3767 & 0.4370 & 0.2512 & 0.1825 & 0.4791 \\
Plane & 105 & 105 & 210 & 144 & 7 & 0.9286 & 0.6869 & 0.9286 & \textbf{1.0000} & 0.9905 & 0.9905 & 0.9905 \\
PowerCons & 180 & 180 & 360 & 144 & 2 & 0.8778 & 0.6292 & 0.8194 & 0.9440 & 0.9333 & \textbf{0.9694} & 0.9444 \\
ProximalPhalanxOutlineAgeGroup & 400 & 205 & 605 & 80 & 3 & 0.6033 & 0.7726 & \textbf{0.8512} & 0.8510 & 0.7785 & 0.7983 & 0.8364 \\
ProximalPhalanxOutlineCorrect & 600 & 291 & 891 & 80 & 2 & 0.6820 & \textbf{0.8028} & 0.7654 & 0.6760 & 0.7348 & 0.7442 & 0.7935 \\
ProximalPhalanxTW & 400 & 205 & 605 & 80 & 6 & 0.6992 & \textbf{0.8742} & 0.7107 & 0.6860 & 0.7636 & 0.8017 & 0.8050 \\
RefrigerationDevices & 375 & 375 & 750 & 720 & 3 & 0.5000 & 0.6895 & 0.6000 & 0.6000 & 0.5933 & 0.6213 & \textbf{0.6960} \\
Rock & 20 & 50 & 70 & 2844 & 4 & 0.5857 & 0.7400 & 0.5000 & \textbf{0.7860} & 0.4000 & 0.7714 & 0.7286 \\
ScreenType & 375 & 375 & 750 & 720 & 3 & 0.3880 & 0.6111 & 0.3200 & 0.4800 & 0.4293 & 0.5400 & \textbf{0.6400} \\
SemgHandGenderCh2 & 300 & 600 & 900 & 1500 & 2 & 0.6122 & 0.5667 & 0.7167 & 0.8110 & 0.7322 & 0.6000 & \textbf{0.8444} \\
SemgHandMovementCh2 & 450 & 450 & 900 & 1500 & 6 & 0.1667 & 0.4069 & 0.4944 & 0.3830 & 0.2533 & \textbf{0.5067} & 0.4389 \\
SemgHandSubjectCh2 & 450 & 450 & 900 & 1500 & 5 & 0.3567 & 0.6802 & 0.6000 & 0.7060 & 0.6133 & 0.2000 & \textbf{0.7667} \\
ShakeGestureWiimoteZ & 50 & 50 & 100 & Vary & 10 & 0.7700 & 0.2251 & 0.3000 & 0.7000 & 0.8800 & \textbf{0.9200} & 0.8700 \\
ShapeletSim & 20 & 180 & 200 & 500 & 2 & \textbf{1.0000} & 0.7200 & 0.4750 & 0.5500 & 0.8850 & 0.5700 & \textbf{1.0000} \\
ShapesAll & 600 & 600 & 1200 & 512 & 60 & 0.3767 & 0.3016 & 0.7750 & 0.7290 & 0.7592 & 0.8467 & \textbf{0.8692} \\
SmallKitchenAppliances & 375 & 375 & 750 & 720 & 3 & 0.4453 & \textbf{0.7789} & 0.6667 & 0.6800 & 0.6253 & 0.6907 & 0.7707 \\
SmoothSubspace & 150 & 150 & 300 & 15 & 3 & 0.9400 & 0.4559 & 0.9333 & 0.8670 & 0.9600 & 0.8333 & \textbf{0.9667} \\
SonyAIBORobotSurface1 & 20 & 601 & 621 & 70 & 2 & 0.9806 & 0.9762 & 0.9920 & \textbf{1.0000} & 0.9903 & 0.9935 & 0.9678 \\
SonyAIBORobotSurface2 & 27 & 953 & 980 & 65 & 2 & 0.9765 & 0.8250 & 0.9694 & 0.8780 & \textbf{0.9878} & 0.9755 & 0.9337 \\
StarLightCurves & 1000 & 8236 & 9236 & 1024 & 3 & 0.9516 & 0.8132 & 0.9784 & 0.9690 & 0.9670 & \textbf{0.9790} & 0.9710 \\
Strawberry & 613 & 370 & 983 & 235 & 2 & 0.6447 & 0.7340 & 0.9188 & 0.7060 & 0.8640 & 0.6450 & \textbf{0.9847} \\
SwedishLeaf & 500 & 625 & 1125 & 128 & 15 & 0.7342 & 0.7901 & 0.8578 & 0.8800 & 0.8791 & \textbf{0.9218} & 0.9049 \\
Symbols & 25 & 995 & 1020 & 398 & 6 & 0.9843 & 0.4800 & \textbf{0.9902} & 0.9710 & 0.9843 & 0.9843 & 0.9853 \\
SyntheticControl & 300 & 300 & 600 & 60 & 6 & 0.9850 & 0.6429 & 0.9500 & 0.9580 & \textbf{0.9867} & 0.9633 & 0.9767 \\
ToeSegmentation1 & 40 & 228 & 268 & 277 & 2 & 0.5185 & 0.4133 & 0.9444 & 0.7410 & 0.9593 & \textbf{0.9626} & 0.9245 \\
ToeSegmentation2 & 36 & 130 & 166 & 343 & 2 & 0.9212 & 0.7500 & 0.7647 & 0.7940 & \textbf{0.9455} & 0.8856 & 0.8374 \\
Trace & 100 & 100 & 200 & 275 & 4 & 0.6550 & 0.6744 & \textbf{1.0000} & 0.9250 & \textbf{1.0000} & 0.9300 & 0.9750 \\
TwoLeadECG & 23 & 1139 & 1162 & 82 & 2 & 0.9957 & 0.4244 & \textbf{1.0000} & \textbf{1.0000} & \textbf{1.0000} & 0.9957 & \textbf{1.0000} \\
TwoPatterns & 1000 & 4000 & 5000 & 128 & 4 & 0.9892 & 0.7367 & \textbf{1.0000} & 0.9840 & 0.9952 & 0.9984 & \textbf{1.0000} \\
UMD & 36 & 144 & 180 & 150 & 3 & 0.6056 & 0.8200 & 0.7778 & 0.9720 & 0.8556 & 0.9056 & \textbf{0.9778} \\
UWaveGestureLibraryAll & 896 & 3582 & 4478 & 945 & 8 & 0.9542 & 0.7500 & 0.9576 & 0.9430 & 0.9685 & \textbf{0.9761} & 0.9283 \\
UWaveGestureLibraryX & 896 & 3582 & 4478 & 315 & 8 & 0.7904 & 0.7133 & 0.7868 & 0.7790 & \textbf{0.8261} & 0.8249 & 0.8106 \\
UWaveGestureLibraryY & 896 & 3582 & 4478 & 315 & 8 & 0.6855 & 0.6627 & 0.6987 & 0.7230 & 0.7315 & 0.7151 & \textbf{0.7620} \\
UWaveGestureLibraryZ & 896 & 3582 & 4478 & 315 & 8 & 0.7462 & \textbf{0.8200} & 0.7801 & 0.7050 & 0.7638 & 0.7867 & 0.7416 \\
Wafer & 1000 & 6164 & 7164 & 152 & 2 & 0.9939 & 0.9566 & 0.9986 & 0.9980 & 0.9955 & 0.9985 & \textbf{1.0000} \\
Wine & 57 & 54 & 111 & 234 & 2 & 0.4909 & \textbf{0.8663} & 0.5652 & 0.4780 & 0.4727 & 0.5308 & 0.6383 \\
WordSynonyms & 267 & 638 & 905 & 270 & 25 & 0.4751 & 0.6345 & 0.4972 & 0.6410 & 0.6939 & 0.6309 & \textbf{0.8011} \\
Worms & 181 & 77 & 258 & 900 & 5 & 0.4615 & \textbf{0.8969} & 0.4808 & 0.5770 & 0.5308 & 0.5624 & 0.8462 \\
WormsTwoClass & 181 & 77 & 258 & 900 & 2 & 0.5846 & \textbf{0.8860} & 0.6538 & 0.6730 & 0.7346 & 0.6474 & 0.8846 \\
Yoga & 300 & 3000 & 3300 & 426 & 2 & 0.7985 & 0.8142 & 0.8530 & 0.7210 & 0.8694 & \textbf{0.9445} & 0.9385 \\
\midrule 
Avg. Acc &  &  &  &  &  & 0.6262 & 0.7213 & 0.7018 & 0.7479 & 0.7375 & 0.7820 & 0.8405 \\
Avg. Rank &  &  &  &  &  & 5.7969 & 4.3711 & 4.2773 & 4.1172 & 3.9375 & 3.1055 & 2.3945 \\
\end{longtable}
\vspace{-1em}
\captionof{table}{UCR dataset information and accuracy comparison.}
\label{tab:ucr_info_performance_long}

\twocolumn
\onecolumn
\footnotesize
\setlength\LTleft{\fill}
\setlength\LTright{\fill}
\setlength{\tabcolsep}{3pt}
\renewcommand{\arraystretch}{1.05}

\begin{longtable}{l|ccccc|ccccc|c}
\toprule
dataset & \#train & \#test & \#total & length & \#class & LTS & ShapeNet & SVP-T & ShapeConv & SBM & Ours \\
\midrule
\endfirsthead

\toprule
dataset & \#train & \#test & \#total & length & \#class & LTS & ShapeNet & SVP-T & ShapeConv & SBM & Ours \\
\midrule
\endhead

\midrule
\multicolumn{12}{r}{\small\itshape Continued on next page}\\
\endfoot

\bottomrule
\endlastfoot

ArticularyWordRecognition & 275 & 300 & 575 & 144 & 25 & 0.9907 & \textbf{0.9933} & 0.9833 & 0.9767 & \textbf{0.9933} & 0.9733 \\
AtrialFibrillation & 15 & 15 & 30 & 640 & 3 & 0.4533 & 0.1250 & 0.2667 & \textbf{0.4667} & 0.4533 & 0.4400 \\
BasicMotions & 40 & 40 & 80 & 100 & 4 & \textbf{1.0000} & \textbf{1.0000} & \textbf{1.0000} & \textbf{1.0000} & \textbf{1.0000} & 0.9750 \\
CharacterTrajectories & 1422 & 1436 & 2858 & 119 & 20 & 0.9677 & 0.9624 & 0.9847 & 0.9819 & 0.9799 & \textbf{0.9882} \\
Cricket & 108 & 72 & 180 & 1197 & 12 & 0.9861 & 0.9444 & \textbf{1.0000} & 0.8611 & 0.9806 & 0.9861 \\
DuckDuckGeese & 50 & 50 & 100 & 270 & 5 & 0.4360 & 0.3600 & 0.4400 & 0.4200 & 0.3840 & \textbf{0.6320} \\
ERing & 30 & 270 & 300 & 65 & 6 & 0.9622 & 0.9259 & 0.8630 & 0.8704 & \textbf{0.9667} & 0.9096 \\
EigenWorms & 128 & 131 & 259 & 17948 & 5 & 0.5710 & 0.5606 & 0.6260 & 0.5649 & 0.6092 & \textbf{0.7542} \\
Epilepsy & 137 & 138 & 275 & 206 & 4 & 0.9870 & 0.8841 & 0.9783 & 0.9058 & \textbf{0.9928} & 0.9754 \\
EthanolConcentration & 261 & 263 & 524 & 1751 & 4 & 0.2935 & 0.2348 & 0.2776 & 0.2738 & \textbf{0.3049} & 0.2814 \\
FaceDetection & 5890 & 3524 & 9414 & 62 & 2 & 0.6598 & 0.5681 & 0.5094 & 0.5738 & \textbf{0.6633} & 0.6560 \\
FingerMovements & 316 & 100 & 416 & 50 & 2 & 0.5640 & 0.4200 & 0.5700 & 0.5900 & 0.6080 & \textbf{0.6100} \\
HandMovementDirection & 160 & 74 & 234 & 400 & 4 & 0.4649 & 0.2432 & 0.2703 & 0.4054 & \textbf{0.4757} & 0.4054 \\
Handwriting & 150 & 580 & 730 & 52 & 26 & 0.1391 & 0.2753 & 0.5000 & 0.1635 & 0.3624 & \textbf{0.6294} \\
Heartbeat & 204 & 205 & 409 & 405 & 2 & 0.7356 & 0.6408 & 0.7610 & \textbf{0.7805} & 0.7366 & 0.7630 \\
InsectWingbeat & 25000 & 25000 & 50000 & 22 & 10 & 0.5272 & 0.2425 & 0.5772 & 0.1653 & 0.4594 & \textbf{0.6829} \\
JapaneseVowels & 270 & 370 & 640 & 29 & 9 & 0.9627 & 0.9351 & 0.9811 & 0.9081 & 0.9632 & \textbf{0.9860} \\
LSST & 2459 & 2466 & 4925 & 36 & 14 & \textbf{0.6437} & 0.5393 & 0.3208 & 0.3491 & 0.6349 & 0.3526 \\
Libras & 180 & 180 & 360 & 45 & 15 & 0.7733 & 0.7444 & 0.8611 & 0.7889 & 0.8589 & \textbf{0.8844} \\
MotorImagery & 278 & 100 & 378 & 3000 & 2 & 0.6640 & 0.4600 & 0.4800 & 0.5900 & \textbf{0.6860} & 0.6000 \\
NATOPS & 180 & 180 & 360 & 51 & 6 & 0.8778 & 0.8111 & 0.9056 & 0.8944 & 0.8789 & \textbf{0.9556} \\
PEMS-SF & 267 & 173 & 440 & 144 & 7 & 0.8301 & 0.8506 & \textbf{0.8555} & 0.8035 & 0.8428 & 0.8220 \\
PenDigits & 7494 & 3498 & 10992 & 8 & 10 & 0.9618 & 0.8668 & 0.9377 & \textbf{0.9794} & 0.9719 & 0.9731 \\
PhonemeSpectra & 3315 & 3353 & 6668 & 217 & 39 & 0.2401 & 0.1795 & 0.1524 & 0.1047 & 0.2535 & \textbf{0.3023} \\
RacketSports & 151 & 152 & 303 & 30 & 4 & 0.8829 & 0.8026 & 0.8684 & 0.8158 & \textbf{0.8974} & 0.8816 \\
SelfRegulationSCP1 & 268 & 293 & 561 & 896 & 2 & 0.8689 & 0.8231 & 0.7065 & \textbf{0.8874} & 0.8642 & 0.8703 \\
SelfRegulationSCP2 & 200 & 180 & 380 & 1152 & 2 & 0.5333 & \textbf{0.5667} & 0.5000 & 0.5611 & 0.5289 & 0.5556 \\
SpokenArabicDigits & 6599 & 2199 & 8798 & 93 & 10 & 0.9928 & 0.9273 & 0.9636 & 0.9345 & 0.9939 & \textbf{0.9982} \\
StandWalkJump & 12 & 15 & 27 & 2500 & 3 & 0.5600 & 0.2500 & 0.3333 & 0.5333 & \textbf{0.5733} & 0.5600 \\
UWaveGestureLibrary & 120 & 320 & 440 & 315 & 8 & \textbf{0.9194} & 0.7812 & 0.8750 & 0.8156 & 0.9125 & 0.8844 \\
\midrule 
Avg. Acc &  &  &  &  &  & 0.7150 & 0.6306 & 0.6783 & 0.6655 & 0.7277 & \textbf{0.7429} \\
Avg. Rank &  &  &  &  &  & 3.1833 & 4.9500 & 3.7000 & 4.0500 & 2.5667 & \textbf{2.5500} \\
\end{longtable}
\vspace{-1em}
\captionof{table}{UEA dataset information and accuracy comparison.}
\label{tab:uea_info_performance_long}

\twocolumn
\onecolumn
\scriptsize
\setlength\LTleft{\fill}
\setlength\LTright{\fill}
\setlength{\tabcolsep}{3pt}
\renewcommand{\arraystretch}{1.05}

\begin{longtable}{l|ccccc|ccccccc}
\toprule
dataset & \#train & \#test & \#total & length & \#class 
& LTS & ShapeNet & SVP-T & ShapeConv & SBM & SoftShape & Ours (Pop.) \\
\midrule
\endfirsthead

\toprule
dataset & \#train & \#test & \#total & length & \#class 
& LTS & ShapeNet & SVP-T & ShapeConv & SBM & SoftShape & Ours (Pop.) \\
\midrule
\endhead

\midrule
\multicolumn{13}{r}{\small\itshape Continued on next page}\\
\endfoot

\bottomrule
\endlastfoot
ArrowHead & 36 & 175 & 211 & 251 & 3 & 0.5762 & \textbf{0.9529} & 0.5116 & 0.7910 & 0.7381 & 0.9051 & 0.8605 \\
CBF & 30 & 900 & 930 & 128 & 3 & 0.9978 & 0.9400 & 0.9946 & \textbf{1.0000} & \textbf{1.0000} & \textbf{1.0000} & 0.9194 \\
CricketX & 390 & 390 & 780 & 300 & 12 & 0.4974 & \textbf{0.9478} & 0.6538 & 0.5900 & 0.6538 & 0.7718 & 0.7244 \\
DistalPhalanxOutlineAgeGroup & 400 & 139 & 539 & 80 & 3 & 0.7370 & \textbf{0.9276} & 0.7315 & 0.7780 & 0.7944 & 0.7736 & 0.9074 \\
DistalPhalanxOutlineCorrect & 600 & 276 & 876 & 80 & 2 & 0.6354 & \textbf{0.9750} & 0.7784 & 0.7840 & 0.7771 & 0.6587 & 0.7898 \\
ECG5000 & 500 & 4500 & 5000 & 140 & 5 & 0.9332 & \textbf{0.9759} & 0.9590 & 0.9490 & 0.9416 & 0.9516 & 0.9560 \\
EOGVerticalSignal & 362 & 362 & 724 & 1250 & 12 & 0.4566 & \textbf{0.7588} & 0.6552 & 0.4550 & 0.5821 & 0.7029 & 0.6207 \\
EthanolLevel & 504 & 500 & 1004 & 1751 & 4 & 0.2537 & \textbf{0.9222} & 0.2736 & 0.7260 & 0.2557 & 0.4152 & 0.6517 \\
Fish & 175 & 175 & 350 & 463 & 7 & 0.5229 & 0.7079 & 0.6286 & 0.8710 & 0.7486 & \textbf{0.9171} & 0.8143 \\
GunPoint & 50 & 150 & 200 & 150 & 2 & 0.7300 & 0.6273 & 0.8000 & 0.9750 & 0.9600 & \textbf{0.9850} & 0.9750 \\
InsectWingbeatSound & 220 & 1980 & 2200 & 256 & 11 & 0.6136 & 0.5404 & 0.5727 & 0.6520 & 0.6791 & \textbf{0.6995} & 0.6500 \\
ItalyPowerDemand & 67 & 1029 & 1096 & 24 & 2 & 0.9616 & 0.6099 & \textbf{0.9727} & 0.9640 & 0.9598 & 0.9316 & 0.8773 \\
MelbournePedestrian & 1194 & 2439 & 3633 & 24 & 10 & 0.6850 & \textbf{0.9725} & 0.8583 & 0.8460 & 0.8459 & 0.7947 & 0.7840 \\
MiddlePhalanxTW & 399 & 154 & 553 & 80 & 6 & 0.6036 & 0.5043 & 0.5946 & 0.6220 & 0.6216 & 0.4549 & \textbf{0.6396} \\
MixedShapesRegularTrain & 500 & 2425 & 2925 & 1024 & 5 & 0.8503 & 0.5337 & 0.9419 & 0.9330 & 0.9032 & 0.8335 & \textbf{0.9487} \\
OSULeaf & 200 & 242 & 442 & 427 & 6 & 0.5910 & 0.4762 & 0.5955 & 0.6740 & 0.7416 & 0.7965 & \textbf{0.8764} \\
Trace & 100 & 100 & 200 & 275 & 4 & 0.6550 & 0.6744 & \textbf{1.0000} & 0.9250 & \textbf{1.0000} & 0.9300 & 0.9250 \\
WordSynonyms & 267 & 638 & 905 & 270 & 25 & 0.4751 & 0.6345 & 0.4972 & 0.6410 & 0.6939 & 0.6309 & \textbf{0.7403} \\
\midrule 
Avg. Acc &  &  &  &  &  & 0.6542 & 0.7601 & 0.7233 & 0.7876 & 0.7720 & 0.7863 & \textbf{0.8145 }\\
Avg. Rank &  &  &  &  &  & 5.9444 & 3.8889 & 4.2778 & 3.4444 & 3.7778 & 3.5000 & \textbf{3.1667} \\
\end{longtable}
\vspace{-1em}
\captionof{table}{UCR18 dataset information and accuracy comparison (population-level shapelets).}
\label{tab:ucr18_info_performance_pop}

\twocolumn

\onecolumn
\begin{table}[H]
\centering
\small
\setlength{\tabcolsep}{4pt}
\renewcommand{\arraystretch}{1.15}
\begin{tabular}{l|ccc|ccc|ccc}
\toprule
 & \multicolumn{3}{c}{TimesNet} & \multicolumn{3}{c}{ModernTCN} & \multicolumn{3}{c}{InceptionTime} \\
\cmidrule(lr){2-4}\cmidrule(lr){5-7}\cmidrule(lr){8-10}
Dataset & Acc & Coverage & Shapelet Num & Acc & Coverage & Shapelet Num & Acc & Coverage & Shapelet Num \\
\midrule
ArrowHead & 0.8810 & 0.5461 & 3.0476 & \textbf{0.9070} & 0.4617 & 3.1628 & 0.9048 & 0.5029 & 3.0714 \\
CBF & \textbf{1.0000} & 0.4466 & 1.0000 & \textbf{1.0000} & 0.4505 & 0.9892 & \textbf{1.0000} & 0.4493 & 2.0430 \\
DistalPhalanxOutlineAgeGroup & 0.8333 & 0.5116 & 1.3704 & \textbf{0.8611} & 0.4659 & 1.8704 & 0.8519 & 0.5440 & 2.9722 \\
DistalPhalanxOutlineCorrect & \textbf{0.8171} & 0.6414 & 2.9714 & 0.7886 & 0.5054 & 1.4743 & 0.8114 & 0.5736 & 2.9657 \\
ECG5000 & 0.9570 & 0.4862 & 2.2390 & 0.9500 & 0.3768 & 2.0960 & \textbf{0.9600} & 0.3938 & 2.5910 \\
EOGVerticalSignal & 0.4759 & 0.4359 & 1.5310 & 0.6276 & 0.4598 & 1.8000 & \textbf{0.7724} & 0.4996 & 1.7931 \\
EthanolLevel & 0.5672 & 0.6212 & 2.0000 & \textbf{0.8905} & 0.6213 & 2.0000 & 0.8600 & 0.2659 & 2.3100 \\
Fish & 0.8000 & 0.6734 & 3.0000 & \textbf{0.8571} & 0.5040 & 2.0000 & \textbf{0.8571} & 0.5043 & 3.0000 \\
GunPoint & 0.9500 & 0.5450 & 1.9750 & 0.9750 & 0.5667 & 2.0000 & \textbf{1.0000} & 0.5458 & 2.0000 \\
InsectWingbeatSound & 0.6409 & 0.4082 & 3.4659 & 0.6409 & 0.3206 & 3.1909 & \textbf{0.6591} & 0.3229 & 3.4659 \\
ItalyPowerDemand & 0.9635 & 0.5672 & 1.0183 & 0.9406 & 0.3721 & 0.7078 & \textbf{0.9772} & 0.5200 & 1.1370 \\
MelbournePedestrian & \textbf{0.8664} & 0.5670 & 1.0014 & 0.8652 & 0.6356 & 1.0014 & 0.8583 & 0.5556 & 1.6259 \\
MiddlePhalanxTW & 0.6273 & 0.5261 & 2.5182 & 0.6306 & 0.5310 & 2.5315 & \textbf{0.6667} & 0.4623 & 2.1622 \\
MixedShapesRegularTrain & 0.9111 & 0.4806 & 5.0393 & 0.9470 & 0.5107 & 5.0530 & \textbf{0.9709} & 0.5017 & 3.0752 \\
OSULeaf & 0.6023 & 0.6875 & 5.6023 & 0.6136 & 0.6498 & 5.9432 & \textbf{0.8409} & 0.5812 & 7.2273 \\
Trace & \textbf{1.0000} & 0.9859 & 1.2250 & 0.9750 & 1.0000 & 1.0000 & 0.9750 & 0.5527 & 1.7500 \\
WordSynonyms & 0.6298 & 0.3950 & 2.6575 & 0.6851 & 0.6108 & 3.2210 & \textbf{0.8011} & 0.3820 & 3.1602 \\
\bottomrule
\end{tabular}
\caption{Predictor architecture ablation results. Best accuracy per dataset is highlighted in bold.}
\label{tab:pred_arch_ablation}
\end{table}
\twocolumn
\onecolumn
\begin{table*}[t]
\centering
\scriptsize
\setlength{\tabcolsep}{2pt}
\renewcommand{\arraystretch}{1.12}

\resizebox{\textwidth}{!}{%
\begin{tabular}{l|cccc|cccc|cccc|cccc}
\toprule
 & \multicolumn{4}{c}{$W=1$} & \multicolumn{4}{c}{$W=2$} & \multicolumn{4}{c}{B\_Spline} & \multicolumn{4}{c}{PELT} \\
\cmidrule(lr){2-5}\cmidrule(lr){6-9}\cmidrule(lr){10-13}\cmidrule(lr){14-17}
Dataset
& Acc & Cov & \#S & Rank
& Acc & Cov & \#S & Rank
& Acc & Cov & \#S & Rank
& Acc & Cov & \#S & Rank \\
\midrule
ArrowHead & 0.7677 & 0.5119 & 3.5936 & 4.0000 & 0.8060 & 0.5833 & 3.3173 & 2.0000 & 0.7869 & 0.5326 & 1.6163 & 3.0000 & \textbf{0.8519} & 0.5440 & 2.9722 & 1.0000 \\
CBF & \textbf{0.9957} & 0.4846 & 12.0602 & 1.0000 & 0.9785 & 0.6252 & 9.7151 & 3.0000 & 0.9903 & 0.5118 & 6.3022 & 2.0000 & 0.9048 & 0.5029 & 3.0714 & 4.0000 \\
CricketX & 0.8231 & 0.5141 & 16.2231 & 3.0000 & 0.7346 & 0.3164 & 12.9460 & 4.0000 & 0.8308 & 0.6043 & 4.8872 & 2.0000 & \textbf{0.8519} & 0.5440 & 2.9630 & 1.0000 \\
DistalPhalanxOutlineAgeGroup & \textbf{0.8163} & 0.5548 & 3.2344 & 1.0000 & 0.7977 & 0.5645 & 3.1691 & 4.0000 & 0.8107 & 0.3267 & 1.3807 & 3.0000 & 0.8114 & 0.5736 & 2.9657 & 2.0000 \\
DistalPhalanxOutlineCorrect & 0.7865 & 0.6255 & 2.8465 & 2.0000 & 0.7683 & 0.4878 & 2.7656 & 4.0000 & \textbf{0.7957} & 0.6651 & 1.9126 & 1.0000 & 0.7724 & 0.4996 & 1.7931 & 3.0000 \\
ECG5000 & 0.9468 & 0.3727 & 5.7886 & 2.0000 & 0.9290 & 0.5017 & 4.3428 & 3.0000 & \textbf{0.9506} & 0.4310 & 2.5948 & 1.0000 & 0.8600 & 0.2659 & 2.3100 & 4.0000 \\
EOGVerticalSignal & 0.5691 & 0.5940 & 5.2389 & 3.0000 & 0.4889 & 0.5206 & 3.3368 & 4.0000 & 0.6823 & 0.6179 & 1.9849 & 2.0000 & \textbf{0.8571} & 0.5043 & 3.0000 & 1.0000 \\
EthanolLevel & 0.4920 & 0.5814 & 2.8291 & 4.0000 & 0.5933 & 0.4829 & 1.8935 & 3.0000 & 0.7561 & 0.5273 & 1.8209 & 2.0000 & \textbf{1.0000} & 0.5458 & 2.0000 & 1.0000 \\
Fish & \textbf{0.9114} & 0.5226 & 2.6400 & 1.0000 & 0.7029 & 0.3449 & 3.0371 & 3.0000 & 0.8914 & 0.5119 & 2.2000 & 2.0000 & 0.6591 & 0.3229 & 3.4659 & 4.0000 \\
GunPoint & 0.9650 & 0.5554 & 1.8850 & 3.5000 & 0.9650 & 0.6046 & 2.0300 & 3.5000 & 0.9750 & 0.3743 & 1.0500 & 2.0000 & \textbf{0.9772} & 0.5200 & 1.1370 & 1.0000 \\
InsectWingbeatSound & 0.6241 & 0.5429 & 7.6386 & 3.0000 & 0.5909 & 0.4388 & 3.6136 & 4.0000 & 0.6577 & 0.5925 & 2.7309 & 2.0000 & \textbf{0.8583} & 0.5556 & 1.6259 & 1.0000 \\
ItalyPowerDemand & \textbf{0.9544} & 0.4069 & 1.8926 & 1.0000 & 0.9142 & 0.5626 & 1.7382 & 2.0000 & 0.8805 & 0.4240 & 1.0037 & 3.0000 & 0.6667 & 0.4623 & 2.1622 & 4.0000 \\
MelbournePedestrian & 0.8560 & 0.4689 & 1.5189 & 2.0000 & 0.7149 & 0.6505 & 1.2103 & 4.0000 & 0.8525 & 0.5564 & 1.2103 & 3.0000 & \textbf{0.9709} & 0.5017 & 3.0752 & 1.0000 \\
MiddlePhalanxTW & 0.6239 & 0.5171 & 3.0272 & 2.0000 & 0.5985 & 0.5660 & 2.7353 & 3.0000 & 0.5785 & 0.5498 & 1.0995 & 4.0000 & \textbf{0.8409} & 0.5812 & 7.2273 & 1.0000 \\
MixedShapesRegularTrain & 0.8379 & 0.4536 & 6.8568 & 4.0000 & 0.9422 & 0.4096 & 3.9528 & 3.0000 & 0.9450 & 0.5471 & 3.8971 & 2.0000 & \textbf{0.9750} & 0.5527 & 1.7500 & 1.0000 \\
OSULeaf & 0.7920 & 0.5543 & 10.5734 & 4.0000 & 0.8507 & 0.3530 & 6.0274 & 2.0000 & \textbf{0.8779} & 0.4614 & 6.6648 & 1.0000 & 0.8011 & 0.3820 & 3.1602 & 3.0000 \\
Trace & 0.9150 & 0.6067 & 3.3000 & 3.0000 & 0.6700 & 0.4980 & 5.6850 & 4.0000 & 0.9200 & 0.3879 & 1.9850 & 2.0000 & \textbf{1.0000} & 0.4493 & 2.0430 & 1.0000 \\
WordSynonyms & 0.7083 & 0.6703 & 7.7326 & 4.0000 & 0.7536 & 0.7521 & 7.9580 & 2.0000 & 0.7370 & 0.4115 & 3.4851 & 3.0000 & \textbf{0.9600} & 0.3938 & 2.5910 & 1.0000 \\
\bottomrule
\end{tabular}%
}
\caption{Ablation study comparing fixed length ($W=1$), average PELT length ($W=2$), B-spline smoothing, and PELT-based segmentation. Best accuracy per dataset is highlighted in bold.}
\label{tab:ablation_l_bspline_pelt}
\end{table*}
\twocolumn
\twocolumn





\end{document}